\documentclass{article}
\usepackage{times}
\usepackage{iclr2027_conference}
\iclrfinalcopy

\usepackage{amsmath,amsfonts,bm}

\def\1{\bm{1}}

\DeclareMathAlphabet{\mathsfit}{\encodingdefault}{\sfdefault}{m}{sl}
\SetMathAlphabet{\mathsfit}{bold}{\encodingdefault}{\sfdefault}{bx}{n}

\usepackage{hyperref}
\hypersetup{hidelinks}
\usepackage{url}
\usepackage{booktabs}
\usepackage{amsfonts}
\usepackage{nicefrac}
\usepackage{microtype}
\usepackage{xcolor}
\usepackage{multirow}
\usepackage{adjustbox}
\usepackage{enumitem}
\usepackage{longtable}
\usepackage{titletoc}
\usepackage{amsmath}
\usepackage{tabularx}
\usepackage{array}
\usepackage{ragged2e}
\usepackage{float}
\usepackage{caption,fvextra}
\title{\mbox{ProcVLM: Learning Procedure-Grounded}\\
Progress Rewards for Robotic Manipulation}
\author{%
Youhe Feng$^{1}$\thanks{Work done during the internship at Zhipu AI.} \quad
Hansen Shi$^{1}$ \quad
Haoyang Li$^{1}$ \quad
Xinlei Guo$^{1}$ \quad
Yang Wang$^{1}$ \\
\textbf{Chengyang Zhang}$^{1}$ \quad
\textbf{Jinkai Zhang}$^{1}$ \quad
\textbf{Xiaohan Zhang}$^{2}$ \quad
\textbf{Jie Tang}$^{3}$ \quad
\textbf{Jing Zhang}$^{1}$\thanks{Corresponding author. \texttt{zhang-jing@ruc.edu.cn}} \\
\textnormal{$^{1}$School of Information, Renmin University of China} \\
\textnormal{$^{2}$Zhipu AI} \\
\textnormal{$^{3}$Department of Computer Science and Technology, Tsinghua University}
}
\date{Preprint}
\begin{document}
\maketitle
\fancyhead{}
\begin{center}\textit{Preprint}\end{center}
\begin{abstract}
Long-horizon robotic manipulation requires dense feedback that reflects how a task advances through its procedural stages, not merely whether the final outcome is successful. 
Existing reward models often rely on trajectory-level success labels or time-based interpolation, which can conflate elapsed time with true task progress and therefore fail to capture unfinished steps, stagnation, and failure states.
We present \textbf{ProcVLM}, a progress-aware vision-language model that learns procedure-grounded progress as a dense reward signal for manipulation.
Rather than deriving progress from terminal outcomes or temporal proxies, ProcVLM grounds progress estimation in procedural structure and intra-stage visual change, and further adopts a reasoning-before-estimation paradigm that infers the remaining atomic actions before estimating task progress.
Specifically, we construct this supervision by synthesizing frame-level subtask-semantic annotations, assigning progress budgets according to subtask structure, and distributing each budget based on intra-subtask visual change.
To train ProcVLM at scale, we build a standardized procedural supervision synthesis pipeline and construct ProcCorpus-60M from 30 embodied datasets with 60M annotated frames, from which we derive ProcVQA for procedure-aware pretraining, with progress estimation as the central task alongside action segmentation and future planning.
Experiments on ProcVQA and reward-model benchmarks show that ProcVLM improves embodied procedural reasoning and yields more discriminative trajectory-internal progress estimates than representative baselines, supporting its use as a dense reward model for downstream reward-guided policy optimization.
\end{abstract}
\section{Introduction}

Recent vision-language-action (VLA) models have substantially improved policy generalization from large-scale robot demonstrations~\citep{brohan2022rt1,brohan2023rt2,openx2023,octo2024,kim2024openvla,black2024pi0,pi05_2025,nvidia2025grootn1,gemini2025robotics,gemini2025robotics15,yang2025instructvla,li2025bridgevla,qu2025spatialvla}.
However, demonstration-centric pretraining is constrained by the cost of collecting diverse robot data and the difficulty of adapting pretrained policies to new tasks, environments, and failure modes. 
This has motivated reward-guided policy improvement paradigms, where VLA policies improve beyond offline demonstrations through autonomous experience, expert corrections, learned reward feedback, or self-improvement signals \citep{pi06_2025,xiao2025pld,fei2025srpo,zhai2025vlac}.
Such paradigms require reward models that provide dense, task-conditioned feedback on task progress, remaining steps, stagnation, and failure.

A central challenge is that existing robot datasets rarely provide dense supervision for procedural progress. 
Most large-scale corpora contain task instructions, observations, actions, and sometimes trajectory-level success labels, but lack explicit annotations of procedural structure, such as subtask boundaries and remaining actions~\citep{openx2023,khazatsky2024droid,walke2023bridgedata}. 
As a result, robotic reward or progress models often resort to indirect supervision from sparse outcomes, time-based interpolation, temporal-difference learning, or comparison- and preference-based signals~\citep{zhai2025vlac,lee2026roboreward,huang2025corft,liang2026robometer,tan2025robodopamine,zhang2024grape}.
However, time is not progress in multi-step manipulation.
Linear time-based labels alone provide limited guidance for understanding semantic action stages, task structure, within-stage advancement, and task completion.
During deployment, a policy may spend many steps retrying a grasp, remain stuck in a local stage, or recover from an earlier failure.
The lack of explicit stage information in linear time-based labels can undermine prediction reliability and limit transfer to new tasks and environments.

Motivated by this observation, we introduce \textbf{ProcVLM}, a progress-aware embodied VLM that predicts dense progress rewards grounded in task procedures.
A transferable progress estimator should (i) identify the current execution stage and subtask boundaries, (ii) recognize atomic actions shared across tasks and environments, and (iii) understand task instructions and plan subsequent actions.
To this end, ProcVLM follows a reasoning-before-estimation paradigm: it first infers the current execution stage and remaining atomic actions, and then predicts a continuous progress score conditioned on this explicit procedural reasoning.
To support this learning process, we construct progress labels by assigning each subtask a duration-adjusted progress budget and distributing that budget according to intra-subtask perceptual change.
This coupling provides procedural context for interpreting stage transitions, unfinished steps, stagnation, and failure states.

To train ProcVLM, we develop a highly efficient and scalable procedural supervision synthesis pipeline for embodied trajectory annotation.
The pipeline decomposes raw manipulation trajectories into procedural subtasks and annotates frame-level cues such as state reasoning, completion status, and remaining actions.
Using this pipeline, we construct \textbf{ProcCorpus-60M} from 30 embodied manipulation datasets, covering 400K trajectories and 60M annotated frames. 
We further convert these annotations into \textbf{ProcVQA}, a 20B-token VLM training corpus centered on task progress estimation and complemented by action segmentation and future planning as auxiliary process-reasoning tasks.
ProcVLM is trained in two stages: large-scale procedure-aware pretraining on ProcVQA to learn general procedural representations, followed by refinement on a curated subset to sharpen subtask alignment and progress estimation.
Empirically, ProcVLM improves procedural understanding on ProcVQA and enables one-shot adaptation on RoboFAC~\citep{ye2026robofac}, yielding more discriminative progress estimates than representative reward-model baselines.
Moreover, ProcVLM improves policy performance on RoboTwin~\citep{chen2025robotwin20} through on-policy RL after few-shot reward adaptation. Its rewards also stabilize fine-tuning on noisy real-robot demonstrations, yielding higher early-stage success scores than supervised fine-tuning. 
Our contributions are:

\begin{itemize}[leftmargin=1.2em, labelsep=0.4em, itemsep=0.2em, topsep=0.2em]

    \item We introduce \textit{ProcVLM}, a progress-aware embodied VLM that predicts dense procedure-grounded progress rewards through reasoning-before-estimation, linking continuous progress estimation to subtask semantics, remaining actions, and within-stage advancement.

    \item We develop a scalable procedural supervision synthesis pipeline for converting raw manipulation trajectories into frame-level subtask-structured annotations. 
    It yields \textit{ProcCorpus-60M}, \textit{ProcVQA}, and a two-stage training pipeline: large-scale pretraining followed by curated refinement.

    \item We systematically evaluate ProcVLM across procedure-aware understanding, reward modeling, cross-task adaptation, and downstream policy optimization. The results show that procedure-grounded progress estimation enables transferable reward modeling and supports stable policy improvement.
\end{itemize}
\section{Related Work}

\textbf{Embodied VLMs for Procedural Reasoning.}
A parallel line of work studies how VLMs can support procedural understanding in robotics beyond direct action prediction. 
Early language-grounded robotics systems use foundation models for high-level planning, affordance grounding, program synthesis, and spatial-constraint reasoning \citep{ahn2022saycan,huang2023inner,liang2023codeaspolicies,singh2023progprompt,huang2023voxposer}.
Models such as PaLM-E, VIMA, and Gato further extend this idea by integrating language, vision, and embodied observations into general-purpose architectures for planning and interaction \citep{driess2023palme,jiang2023vima,reed2022gato}.
Robotics-oriented supervision and reasoning models further advance grounded long-horizon understanding, with RoboVQA providing robotics video-text supervision, RoboBrain unifying planning, affordance perception, and trajectory prediction, and PROGRESSLM evaluating task-progress reasoning in VLMs \citep{sermanet2023robovqa,ji2025robobrain,zhang2026progresslm}.
Recent embodied reasoning models further extend VLMs toward spatial grounding, keypoint reasoning, task decomposition, and reinforced embodied reasoning, as represented by ReKep, Gemini Robotics-ER, and related systems \citep{gemini2025robotics,gemini2025robotics15,geminirobotics2025er,ren2024robopoint,liu2024moka,huang2024rekep,yuan2025embodiedr1}.

\textbf{Reward Modeling for Manipulation.}
Earlier work addresses sparse rewards through relabeling and value learning \citep{andrychowicz2017her,huang2025corft}, or foundation-model feedback \citep{zhang2025rewind,lee2026roboreward,chen2026topreward}.
Other frameworks use VLM/LLM feedback, VLA critics, or policy-internal signals for policy improvement \citep{ma2024gvl,wu2026lrm,yong2026vllr,zhai2025vlac,zhang2024grape,lu2025vlarl}.
For dense rewards, Robometer combines frame-level progress/success prediction with trajectory-comparison preferences, while RoboDopamine uses step-wise reward discretization and multi-view fusion \citep{liang2026robometer,tan2025robodopamine}.
More closely related, SARM and SARM2 learn stage-aware rewards, but do not report a dedicated evaluation of reward transfer to held-out environments \citep{sarm2025,sarm22026}.
PRIMO R1 elicits progress reasoning through GRPO, but does not report downstream reward-guided robot policy optimization \citep{primor12026}.
R$^2$VLM uses recurrent reasoning for progress estimation on ALFRED and Ego4D \citep{r2vlm2026}, while DenseReward is trained on a predominantly simulated corpus \citep{densereward2026}.
Complementary approaches use retrieval-based label correction, retry preferences, or transition alignment \citep{urvc2026,retvl2026,dream2reward2026}.
Together, these advances leave large-scale dense procedural reasoning supervision for transferable reward estimation underexplored; we study this direction using 400K trajectories spanning 30 real-robot and simulated datasets.
\section{Methods}

\begin{figure}
    \centering
    \includegraphics[width=\linewidth]{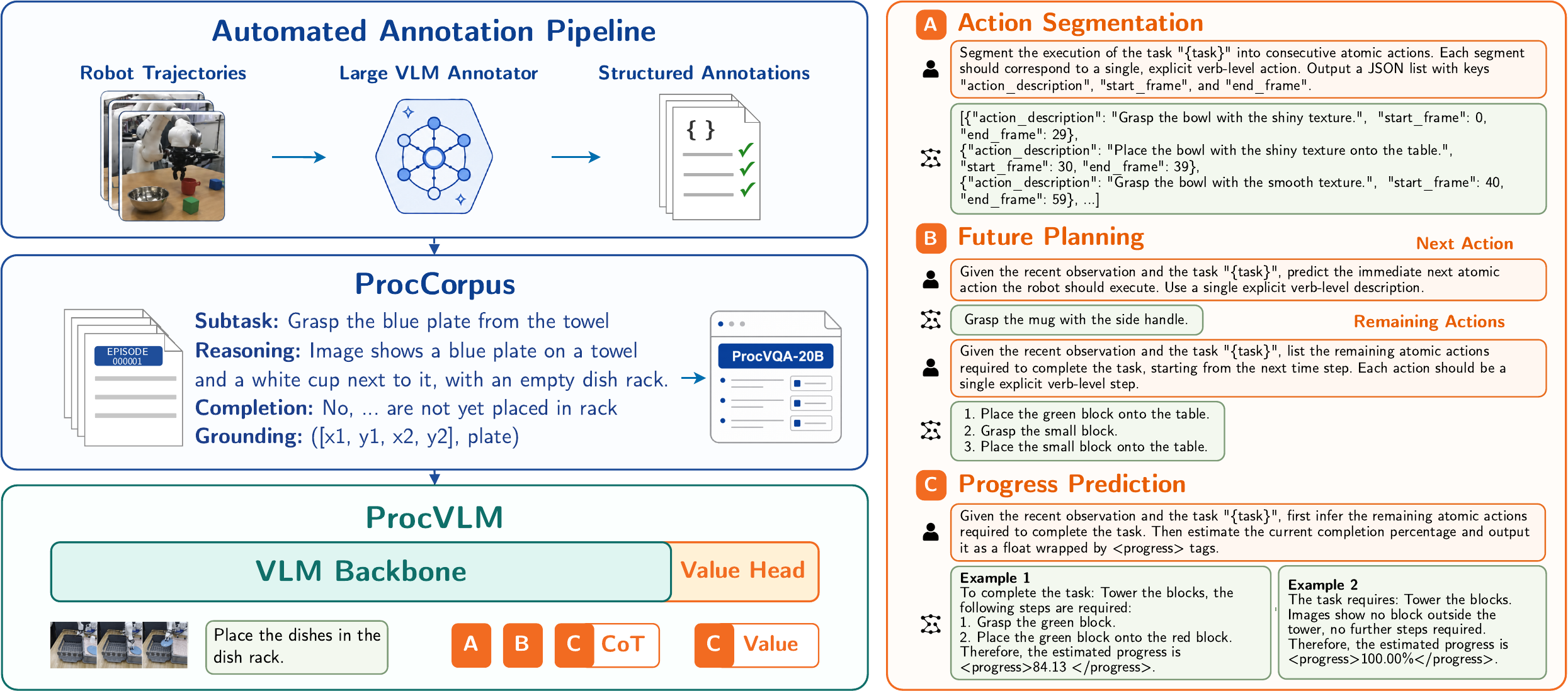}
    \caption{Overview of ProcVLM. A large VLM annotates robot trajectories to form ProcCorpus-60M, which is converted into ProcVQA for action segmentation, future planning, and task progress prediction. Training on ProcVQA enables ProcVLM to provide procedure-grounded progress rewards for downstream policy optimization.}
    \label{fig:procvlm_overview}
\end{figure}

\subsection{Procedural Supervision Synthesis}
\label{sec:pipeline}

Large-scale embodied manipulation datasets contain diverse task executions, but their supervision is usually limited to task instructions, action trajectories, or coarse episode-level outcomes \citep{openx2023,khazatsky2024droid,walke2023bridgedata}. 
To learn procedure-grounded progress rewards, we develop a scalable procedural supervision synthesis pipeline that uses large vision-language models as automatic annotators \citep{bai2025qwen3vl,wang2025internvl35} to convert raw trajectories into frame-wise annotations of subtask stages, completion states, and remaining actions, forming \textit{ProcCorpus-60M}.

\subsubsection{Hierarchical Annotation Pipeline}
\begin{figure}
    \centering
    \includegraphics[width=0.95\linewidth]{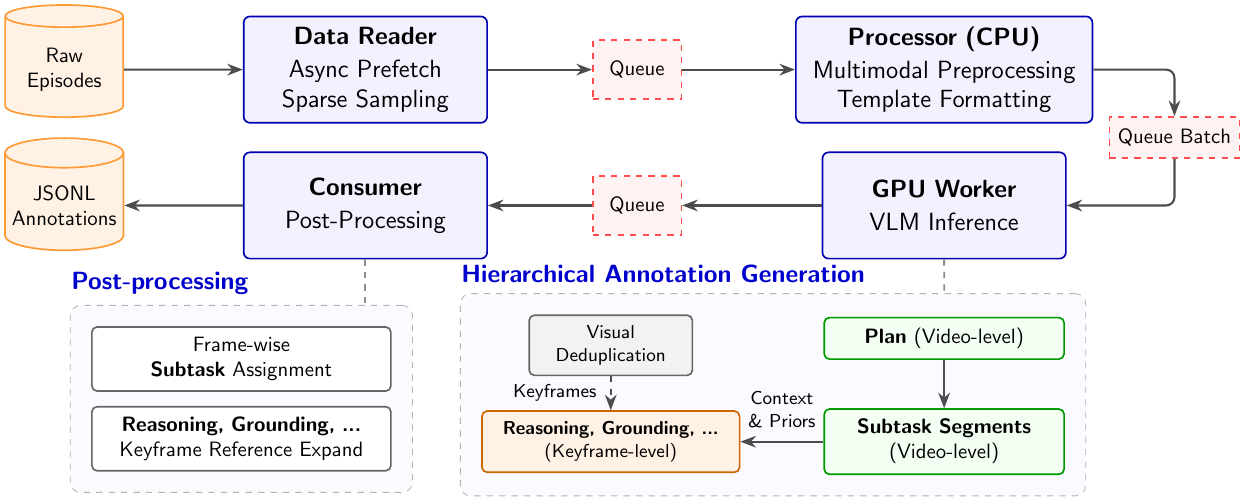}
    \caption{Procedural supervision synthesis pipeline. Asynchronous data reading, multimodal preprocessing, hierarchical VLM annotation, and post-processing convert robot episodes into frame-wise subtask labels and procedural reasoning.}

    \label{fig:annotation_pipeline}
\end{figure}

Figure 2 illustrates our asynchronous annotation pipeline, which decouples data reading, multimodal preprocessing, VLM inference, and post-processing into separate workers connected by queues. 
Annotator and runtime details are provided in Appendix A.

The core annotation process follows a hierarchical, global-to-local design. 
For each episode, we first query the VLM on the complete video to infer a high-level task plan and localize the temporal span of each candidate subtask. 
These video-level segments define a shared subtask structure and are expanded into frame-wise assignments.
Using the resulting task state and remaining subtasks as context, we generate concise visual reasoning on visually deduplicated keyframes.
During post-processing, visually redundant neighboring frames reuse nearby keyframe reasoning, which preserves dense supervision with fewer annotation queries.

\subsubsection{Data Scaling and Filtering}
\label{sec:data_scaling}

We apply the pipeline to real-robot datasets including DROID, BridgeData V2, Fractal, RH20T, Table30, and selected subsets of OXE \citep{khazatsky2024droid,walke2023bridgedata,brohan2022rt1,fang2023rh20t,yakefu2025robochallenge,openx2023}, and simulation datasets including LIBERO, RoboTwin 2.0, and GR00T-Teleop-Sim \citep{liu2023libero,chen2025robotwin20,nvidia2025grootn1}, after screening for visual quality and instruction clarity (Appendix~\ref{app:dataset_details}). The resulting ProcCorpus-60M contains about 400K trajectories and over 60M annotated frames. We fully annotate RoboTwin but exclude it from pretraining and refinement to ensure reliable evaluation.

Each annotated sample contains \textit{(1) task-centric scene reasoning}, including a concise description of task-relevant visual evidence and a completion-state assignment; \textit{(2) subtask annotations}, including the current subtask and the trajectory-level subtask structure; and \textit{(3) optional target grounding annotations} as 2D bounding boxes. Together, they link visual observations to progress, remaining steps, and manipulation targets. For refinement, we manually select about 15K trajectories whose subtask segmentation closely matches the original videos, reducing annotation noise while preserving task diversity, to further improve the model's subtask alignment beyond first-stage pretraining.

Shared trajectory-level subtask structure anchors local annotations, supporting consistent supervision for procedure-grounded progress reward learning. Asynchronous preprocessing and input buffering overlap CPU-side preparation with GPU-side inference, mitigating preprocessing-induced GPU idle time that can arise in native vLLM serving \citep{kwon2023efficient} under image-heavy workloads, and supporting throughput of up to 4M keyframes per day on 8 H100 GPUs under our profiling setting.

\subsection{Learning Procedure-Grounded Progress Rewards}
\label{sec:procvlm_modeling}

Building on ProcCorpus-60M, we introduce \textit{ProcVLM}, an embodied VLM for learning procedure-grounded progress rewards. 
Rather than assigning progress labels through linear time interpolation, ProcVLM learns from subtask-derived targets and conditions progress estimation on explicit reasoning about the current execution stage and remaining steps.
Since ProcCorpus-60M provides structured annotations rather than ready-to-use training samples, we convert them into procedure-aware VQA tasks and jointly train ProcVLM for textual subtask reasoning and continuous progress estimation.

\subsubsection{Procedure-Defined Progress Targets}

To derive continuous progress labels from subtask annotations, we combine subtask-level structure with intra-subtask visual motion. 
Let $T$ denote the trajectory length, $K$ the number of valid subtasks, $k(t)$ the subtask index at time $t$, and $[s_k,e_k]$ the temporal span of the $k$-th subtask. 
We define progress $p(t)$ as the normalized accumulation of local visual change weighted by subtask duration:
\begingroup
\thinmuskip=2mu
\medmuskip=2mu
\thickmuskip=3mu
\begin{equation}
\label{eq:procedure_progress}
p(t)=
\frac{\sum_{0}^{t} w(\tau) r(\tau)}
{\sum_{0}^{T} w(\tau) r(\tau)},
\,
w(\tau)
=
\operatorname{clip}\left(
\frac{K\left(e_{k(\tau)}-s_{k(\tau)}\right)}{T},
0.75,1.25
\right),
r(\tau)
=
\frac{\|\dot{\phi}(\tau)\|}
{\sum_{s_{k(\tau)}}^{e_{k(\tau)}} \|\dot{\phi}(u)\|}.
\end{equation}
\endgroup
Here $\phi(\cdot)$ denotes the visual representation of the observation. 
The weight $w(\tau)$ assigns each subtask a global progress budget according to its relative duration, while the clipping range serves as a soft anchor around an equal-subtask prior, preventing unusually long or short subtasks from dominating or vanishing. 
Since $r(\tau)$ is normalized within each subtask, it distributes this budget over time according to local visual changes. 
In implementation, we approximate $\|\dot{\phi}(\tau)\|$ with adjacent-frame perceptual differences and add a small numerical stabilizer for each frame, yielding a lightweight progress label that preserves subtask structure without reducing the target to linear interpolation over time.

\subsubsection{Procedure-aware Pretraining}
\label{sec:procvlm_pretraining}

To turn the frame-wise annotations in ProcCorpus-60M into learnable supervision, we construct a procedure-aware VQA training set, termed \textit{ProcVQA}, and formulate pretraining as multi-task VQA learning over robot observations and task instructions. 
Each training sample consists of a task instruction, one or more visual observations sampled from the trajectory, and a target response derived from the synthesized annotations. 
As illustrated in Figure~\ref{fig:procvlm_overview}, ProcVQA is centered on task progress estimation and further incorporates two auxiliary process-reasoning tasks to help the model understand task procedures beyond scalar progress prediction. 
Additional details of ProcVQA construction are provided in Appendix~\ref{app:procvqa}.

\textbf{Subtask-structured progress prediction.}
For progress prediction, we supervise the model to output the remaining atomic subtasks before estimating task completion. 
The final response ends with a continuous completion value wrapped by a structured progress tag: \texttt{<progress>p\%</progress>}. 
This format grounds progress estimation in both visual evidence and explicit task structure, while also providing reasoning-formatted supervision for later adaptation.

\textbf{Action segmentation.}
Given a task execution video and its task instruction, the model predicts a sequence of atomic subtask segments with semantic labels and temporal boundaries. 
This auxiliary task supervises the model to identify stage transitions in long-horizon manipulation and builds temporal understanding of task procedures.

\textbf{Future planning.}
Given recent observations and task instruction, the model predicts the executable subtasks required in subsequent steps. 
This auxiliary task focuses on connecting the current manipulation state with future procedural actions, enabling the model to recover the remaining task structure needed for planning and failure recovery.

\subsubsection{ProcVLM Architecture and Training Objectives}
\label{sec:training_objectives}

ProcVLM is built on a compact vision-language backbone initialized from Qwen3-VL-2B-Instruct \citep{bai2025qwen3vl}. 
The backbone takes task instructions and visual observations as input, and generates task-specific textual responses through the standard autoregressive language modeling head. 
This branch supports action segmentation, future planning, and reasoning-formatted progress prediction.

To enable continuous progress prediction, we attach a progress value head on top of the shared VLM representations. 
For progress-estimation samples, this branch is activated to regress a scalar completion score from contextual hidden states produced by the backbone. 
Specifically, the value head applies attention pooling over relevant multimodal and textual representations before predicting the progress value. 
This gated design preserves the original language generation pathway while adding a dedicated continuous regression route for progress estimation, helping mitigate the tendency of token-based numerical prediction to collapse into coarse or quantized anchors~\citep{wang2025softlabeling,ma2026regression,golkar2023xval,zhang2026progresslm}. 
By sharing the backbone with textual subtask reasoning, the value head grounds its prediction in structured visual-language representations rather than isolated numeric cues.

ProcVLM is trained with a joint objective that couples text generation with continuous progress regression. 
Across all three task families, we apply the standard autoregressive language modeling loss to the supervised textual response. 
This also includes progress-estimation samples: their reasoning context and formatted answer are learned through the language modeling objective, while the scalar progress value parsed from the target response's \texttt{<progress>} tag provides additional supervision for the value head.

For samples with progress supervision, the value branch predicts a continuous progress score $\hat{p}$, which is optimized against the ground-truth completion percentage $p$:
\[
\mathcal{L}_{\mathrm{value}} = \ell_{\mathrm{reg}}(\hat{p}, p),
\]
where $\ell_{\mathrm{reg}}$ denotes the regression loss for continuous progress prediction. 
The final training objective is
\begin{equation}
\label{eq:training_objective}
\mathcal{L}
=
\mathcal{L}_{\mathrm{LM}}
+
\lambda \cdot \mathbb{I}_{\mathrm{prog}} \mathcal{L}_{\mathrm{value}},
\end{equation}
where $\mathcal{L}_{\mathrm{LM}}$ is the standard autoregressive language modeling loss, $\mathbb{I}_{\mathrm{prog}}$ indicates whether the sample contains progress supervision, and $\lambda$ controls the weight of the value loss.

We train ProcVLM in two stages. 
The first stage uses the ProcVQA training split with approximately 20B tokens, while the second stage refines the model on a 2.8B-token curated set built from human-selected data to improve subtask alignment, procedural reasoning, and progress estimation. 
Further details on model configuration and training are provided in Appendices~\ref{app:procvlm_setting} and~\ref{app:procvlm_training}.
\section{Experiments}

We evaluate ProcVLM along three questions. 
\textit{(Q1) Does ProcVLM improve embodied procedural understanding?} 
We evaluate this on ProcVQA, covering action segmentation, future planning, and task progress estimation. 
\textit{(Q2) Can ProcVLM serve as a generalizable progress reward model?} 
We compare it with representative robotic reward models, test one-shot adaptation on RoboFAC, and conduct ablations on procedure-aware pretraining and reasoning-formatted supervision. 
\textit{(Q3) Can ProcVLM improve downstream policy learning?} 
We use ProcVLM as a reward model for on-policy RL in RoboTwin simulation and reward-guided fine-tuning on a real robot.

\subsection{Embodied Procedural Understanding}
\label{sec:vlm_exp}

\textbf{Setup.}
We evaluate ProcVLM on ProcVQA, a human-selected subtask-based VQA benchmark for embodied procedural understanding. ProcVQA includes three tasks: action segmentation, future planning, and task progress estimation. 
The ID split is derived from training-domain datasets such as DROID, Bridge, Table30, and OXE, while the OOD split is constructed from unseen RoboTwin tasks.
We compare ProcVLM with mainstream VLMs, including GPT-5.4, Gemini 3.1 Pro, Qwen3VL (235B-A22B), and Qwen3.5 (27B). We report BF1@5/mMAE for action segmentation \citep{ishikawa2021asrf}, human-evaluated Success for future planning, and VOC/EPR@50 for progress estimation. Metric definitions are provided in Appendix~\ref{app:procvqa_metrics}, with EPR defined in Eq.~\eqref{eq:effective_progress_resolution}.

\begin{table}[h]
  \caption{Embodied procedural understanding evaluation on ProcVQA. We compare ProcVLM with mainstream VLMs on ID and OOD splits across action segmentation, future planning, and task progress estimation.}
  \label{tab:id_ood}
  \centering
  \small
  \begin{tabular}{l lccccc}
    \toprule
    & & \multicolumn{2}{c}{Action Segmentation} & Future Planning & \multicolumn{2}{c}{Progress Estimation} \\
    \cmidrule(r){3-4} \cmidrule(r){6-7}
    Dataset & Name & BF1@5 $\uparrow$ & mMAE $\downarrow$ & Success $\uparrow$ & VOC $\uparrow$ & EPR@50 $\uparrow$ \\
    \midrule

    \multirow{5}{*}{ID}
    & Qwen3VL & 0.3945 & 0.6033 & 0.6034 & 0.4420 & 4.4645 \\
    & Qwen3.5 & 0.4296 & 0.7247 & 0.7931 & 0.5475 & 4.8741 \\
    & Gemini 3.1 Pro & 0.4698 & 0.6578 & 0.7413 & 0.5087 & 5.3725 \\
    & GPT-5.4 & 0.5221 & \textbf{0.4864} & 0.5517 & 0.5229 & 4.8058 \\
    & ProcVLM (Ours) & \textbf{0.6924} & 0.6758 & \textbf{0.8103} & \textbf{0.8058} & \textbf{5.7489} \\

    \midrule

    \multirow{5}{*}{OOD}
    & Qwen3VL & 0.4330 & 0.6263 & 0.6896 & 0.5361 & 4.9178 \\
    & Qwen3.5 & 0.4185 & \textbf{0.5131} & 0.7758 & 0.6397 & 5.0932 \\
    & Gemini 3.1 Pro & 0.4584 & 0.6015 & 0.7241 & 0.5987 & 6.0399 \\
    & GPT-5.4 & 0.4941 & 0.5158 & 0.5862 & 0.6553 & 5.2751 \\
    & ProcVLM (Ours) & \textbf{0.5802} & 0.5579 & \textbf{0.8448} & \textbf{0.7282} & \textbf{6.1152} \\

    \bottomrule
  \end{tabular}
\end{table}

\begin{table}
  \small
  \caption{Zero-shot comparison with robotic reward models on the ProcVQA-OOD progress estimation subset.}
  \label{tab:reward_bench}
  \centering
  \begin{tabular}{l lcc@{\hspace{0.8em}}c}
    \toprule
    Dataset & Metric & ProcVLM-2B & Robometer-4B & RoboDopamine-4B \\
    \midrule
    \multirow{2}{*}{ProcVQA-OOD}
    & VOC $\uparrow$ & \textbf{0.7282} & 0.5296 & \underline{0.7156} \\
    & EPR@50 $\uparrow$ & \underline{6.1152} & \textbf{6.7212} & 4.5693 \\
    \bottomrule
  \end{tabular}
\end{table}

\textbf{Findings.} Table~\ref{tab:id_ood} shows that ProcVLM achieves the strongest overall performance on both ID and OOD splits. It obtains the best BF1@5, Success, VOC, and EPR@50, showing consistent gains in action segmentation, future planning, and progress estimation. Although its mMAE is not the lowest, ProcVLM remains comparable on this auxiliary localization metric while clearly improving the primary segmentation metric.

\subsection{Generalizable Progress Reward Evaluation}\label{sec:progress_reward_eval}

\subsubsection{Zero-Shot Comparison with Robotic Reward Models}
\label{sec:reward_model_exp}

\textbf{Baselines and setup.}
We compare ProcVLM with Robometer and RoboDopamine on ProcVQA-OOD under a zero-shot setting.
\emph{Robometer} is trained on RBM-1M, a reward-learning dataset with over one million trajectories, which is larger in trajectory scale than our 400K-trajectory ProcCorpus-60M. 
It combines frame-level progress/success and trajectory-preference supervision without explicit reasoning over current and remaining subtasks. 
\emph{RoboDopamine} predicts relative progress hops from multi-view comparisons of initial, goal, before, and after states through step-wise reward discretization and multi-perspective fusion.
Our primary evaluation uses shuffled local-window queries, with short frame windows evaluated independently without chronological ordering.
For RoboDopamine, we prepend a blank image as the before-state anchor; we acknowledge that this deviates from its intended input format and may limit its performance.
EPR@50 is an auxiliary measure of progress-score resolution; it does not assess progress ordering, which is measured by VOC. 
Input construction and progress-score extraction are detailed in Appendix~\ref{app:baseline_details}.

\textbf{Findings.}
Table~\ref{tab:reward_bench} shows that ProcVLM achieves the best VOC under the shuffled local-window setting on the OOD split of ProcVQA. 
Robometer attains higher EPR@50, likely due to its carefully designed progress-regression head, but its lower VOC indicates weaker trajectory-internal progress ordering under zero-shot evaluation.
RoboDopamine remains relatively stable with a blank contrastive anchor and shuffled VQA-window evaluation, but still lags behind ProcVLM in VOC. Its lower EPR@50 is consistent with the difficulty of representing continuous progress through discrete language tokens, which can favor coarse value anchors~\citep{golkar2023xval,wang2025softlabeling,zausinger2024regress}.

\subsubsection{One-Shot Generalization}
\label{sec:robofac_exp}

\textbf{Setup and metrics.} We evaluate one-shot reward-model generalization on successful and failed executions from the real-robot subset of RoboFAC, a failure-centric robotic VQA benchmark~\citep{ye2026robofac}. The 1-shot (Succ.) setting provides one successful demonstration per task, while 1-shot (Succ. + Fail.) additionally includes one demonstration for each task-specific failure type. Since one-shot demonstrations do not cover all camera views, the test set may contain unseen viewpoints, making the setting closer to real deployment.
We report VOC$_{\mathrm{succ}}$ for temporal ordering within successful trajectories, MAE$_{\mathrm{fail}}$ (Eq.~\eqref{eq:mae_fail}) for fault localization on failed trajectories, MCC for success detection, and full-trajectory inference latency.
Details are in Appendix~\ref{app:robofac_details}.

\begin{table}[t]
  \caption{One-shot generalization on the real-robot subset of RoboFAC. We compare zero-shot and one-shot settings, where 1-shot (Succ.) provides one successful demonstration per task and 1-shot (Succ. + Fail.) additionally provides one demonstration for each task-specific failure type.}
  \label{tab:robofac_bench}
  \centering
  \small
  \begin{tabular}{l l cccc}
    \toprule
    Setting & Name & VOC$_{\mathrm{succ}}$ $\uparrow$ & MAE$_{\mathrm{fail}}$ $\downarrow$ & MCC $\uparrow$ & Latency (s) $\downarrow$ \\
    \midrule
    \multirow{4}{*}{Zero-shot} 
      & Qwen3.5-27B & 0.6191 & 0.1267 & 0.5678 & 1188 \\
      & RoboDopamine-8B & 0.9095 & 0.2268 & 0.5506 & 157 \\
      & Robometer-4B & 0.4129 & \underline{0.1218} & 0.4516 & 139 \\
      & ProcVLM-2B & 0.4920 & 0.2001 & 0.6665 & \textbf{50} \\
    \midrule
    \multirow{2}{*}{1-shot (Succ.)} 
      & Robometer-4B & 0.4094 & 0.1222 & 0.4435 & 136 \\ 
      & ProcVLM-2B & \underline{0.9137} & 0.1241 & \underline{0.7918} & 81 \\
    \midrule
    \multirow{2}{*}{1-shot (Succ. + Fail.)} 
      & Robometer-4B & 0.4040 & 0.1221 & 0.4488 & 144 \\
      & ProcVLM-2B & \textbf{0.9301} & \textbf{0.1187} & \textbf{0.8053} & \underline{80} \\
    \bottomrule
  \end{tabular}
\end{table}

\textbf{Fast one-shot adaptation.} While zero-shot ProcVLM does not achieve the best progress ordering, adding a single successful trajectory per task improves VOC$_{\mathrm{succ}}$ by 85.7\%, increases MCC by 18.8\%, and reduces MAE$_{\mathrm{fail}}$ by 38.0\% (Table~\ref{tab:robofac_bench}). Although the VOC$_{\mathrm{succ}}$ gain may partly reflect adaptation to the successful demonstration, the simultaneous improvements in MAE$_{\mathrm{fail}}$ and MCC indicate that ProcVLM's adaptation extends beyond progress ordering on the demonstrated success case.
In contrast, Robometer shows little improvement under the same one-shot setting, suggesting that preference supervision is less effective for few-demonstration adaptation. ProcVLM also maintains substantially lower inference latency than larger reward-model baselines, benefiting from its more compact model size.

\textbf{Robust success detection and fault localization.} Zero-shot ProcVLM achieves the best MCC (0.6665), which rises to the best overall MCC of 0.8053 with successful and failed demonstrations. For fault localization, one successful demonstration brings MAE$_{\mathrm{fail}}$ close to Robometer despite its larger scale and more extensive reward pretraining; adding failure demonstrations further improves MAE$_{\mathrm{fail}}$ to 0.1187.

\subsubsection{Ablation Studies}

\textbf{Ablation setup.}
We conduct the ablation study under the RoboFAC 1-shot (Succ.) setting described in Section~\ref{sec:robofac_exp}. Variants without ProcVLM pretraining are initialized from Qwen3-VL-2B-Instruct, the same backbone used before our procedure-aware pretraining.

\begin{table}[htbp]
  \caption{Ablation study on RoboFAC 1-shot adaptation. \textit{PT} denotes ProcVLM pretraining and \textit{Rsn} denotes reasoning-formatted supervision. $\Delta$ reports relative performance changes from ProcVLM, where negative values indicate degradation after accounting for metric direction.}
  \label{tab:ablation}
  \centering
  \small
  \begin{tabular}{l c c c c c c c c}
    \toprule
    \multirow{2}{*}{Variant} & \multirow{2}{*}{PT} & \multirow{2}{*}{Rsn.}
    & \multicolumn{2}{c}{VOC$_{\mathrm{succ}}$ $\uparrow$}
    & \multicolumn{2}{c}{MAE$_{\mathrm{fail}}$ $\downarrow$}
    & \multicolumn{2}{c}{MCC $\uparrow$} \\
    \cmidrule(lr){4-5} \cmidrule(lr){6-7} \cmidrule(lr){8-9}
    & & & Score & $\Delta$ & Score & $\Delta$ & Score & $\Delta$ \\
    \midrule
    ProcVLM
      & \checkmark & \checkmark
      & \textbf{0.9137} & --
      & \textbf{0.1241} & --
      & \textbf{0.7918} & -- \\
    w/o Reasoning
      & \checkmark & --
      & 0.9040 & $-1.1\%$
      & 0.1284 & $-3.5\%$
      & 0.7598 & $-4.0\%$ \\
    w/o Pretrain
      & -- & \checkmark
      & 0.8281 & $-9.4\%$
      & 0.2343 & $-88.8\%$
      & 0.5949 & $-24.9\%$ \\
    Base
      & -- & --
      & 0.8795 & $-3.7\%$
      & 0.1985 & $-60.0\%$
      & 0.6854 & $-13.4\%$ \\
    \bottomrule
  \end{tabular}
\end{table}

\textbf{Procedure-aware pretraining enables one-shot transfer.}
Removing ProcVLM pretraining causes the largest degradation, increasing MAE$_{\mathrm{fail}}$ by 88.8\% and reducing MCC by 24.9\% (Table~\ref{tab:ablation}). This indicates that large-scale procedure-aware pretraining provides transferable representations for task progress perception, enabling  the model to adapt to unseen RoboFAC tasks from only one successful demonstration.

\textbf{Subtask-structured reasoning improves progress alignment.}
Removing reasoning-formatted supervision degrades all metrics, though less than removing pretraining, suggesting that reasoning helps align low-level subtask cues with downstream progress labels, improving one-shot generalization. 

The Base variant further supports this conclusion, which retains reasonable progress ordering on successful trajectories after one-shot adaptation but performs worse on unseen failure cases (MAE$_{\mathrm{fail}}$) and success detection (MCC), highlighting the \textbf{complementary roles of pretraining and reasoning}. Appendix~\ref{app:additional_analyses} provides an annotation quality audit, a value head ablation, and qualitative reward modeling cases.

\subsection{Reward-guided Policy Optimization}
\label{sec:rft_exp}
\textbf{On-policy RL on RoboTwin.}
We evaluate ProcVLM, Robometer, and RoboDopamine on two RoboTwin tasks: Place Container Plate and Move Can Pot.
Following the recommended adaptation protocols, we manually stage-annotate two of the 50 demonstrations per task to generate approximately 300 training samples for ProcVLM and 2,000 for RoboDopamine.
Robometer uses 50 demonstrations without manual annotations, with five held out for validation.
We then fine-tune each reward model on its corresponding adaptation data using its recommended configuration.
The adapted reward models provide dense action-chunk rewards for OpenVLA-OFT~\citep{kim2025fine} training with GRPO~\citep{shao2024deepseekmath} in RLinf~\citep{yu2026rlinf}.
Within each task, all methods share the SFT initialization, policy training/evaluation settings, and 480 trajectories per update. We compare success rates against active time on both tasks. Adaptation and evaluation details are in Appendix~\ref{app:robotwin_sim_setup}.

\textbf{Real-robot RFT.}
We conduct real-robot experiments on a stack-bowls task with a JAKA robot.
We perform reward fine-tuning (RFT) with Evo-RL~\citep{evorl2026}, using ProcVLM's progress predictions as dense rewards to derive binary advantage labels (positive or negative) for advantage-conditioned policy training.
Using $\pi_{0.5}$~\citep{pi05_2025} as the base policy, we compare standard supervised fine-tuning (SFT) with RFT using the same policy initialization, batch size, and training set of 50 teleoperated demonstrations.
Further details are provided in Appendix~\ref{app:jaka_real_robot}.

\begin{figure}[htbp]
\centering
\begin{minipage}[t]{0.71\linewidth}
\vspace{0pt}
\centering
\includegraphics[width=\linewidth]{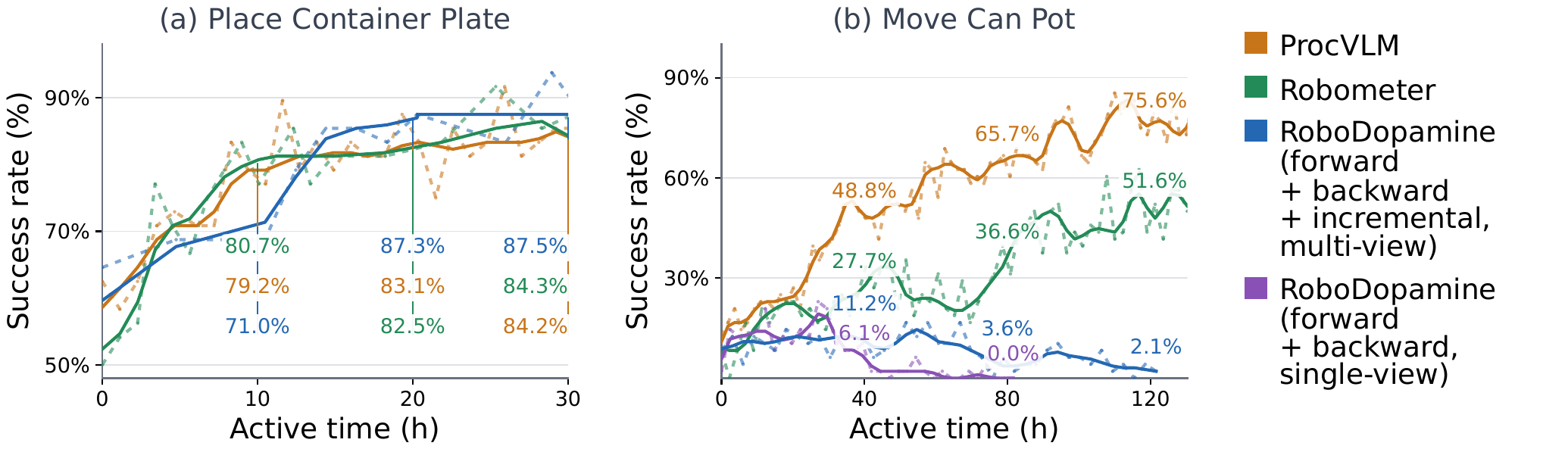}
\caption{On-policy GRPO on RoboTwin.}
\label{fig:robotwin_onpolicy}
\end{minipage}\hfill
\begin{minipage}[t]{0.29\linewidth}
\vspace{0pt}
\centering
\captionsetup{type=table,justification=raggedright,singlelinecheck=false}
\caption{Soft success on the real-robot Stack Bowls task (\%, $\uparrow$).
$\Delta$: gain over SFT in percentage points.}
\label{tab:rft_results}
\footnotesize
\setlength{\tabcolsep}{2pt}
\begin{tabular*}{\linewidth}{@{\extracolsep{\fill}}lccc@{}}
\toprule
Steps & SFT & RFT & $\Delta$ \\
\midrule
5k & 37.5 & 62.5 & +25.0 \\
10k & 70.8 & 83.3 & +12.5 \\
\bottomrule
\end{tabular*}
\end{minipage}
\end{figure}

\textbf{Findings.}
\textit{On the RoboTwin benchmark,} ProcVLM and Robometer achieve comparable success rates on Place Container Plate, where SFT success is relatively high, while RoboDopamine attains slightly higher success near convergence.
On the more challenging Move Can Pot task, where SFT success is low, ProcVLM achieves the highest final success rate, followed by Robometer.
Both RoboDopamine variants remain near the low SFT baseline, suggesting limited transfer to this task under the same settings for fine-tuning the reward model (Figure~\ref{fig:robotwin_onpolicy}).
These results further support the effectiveness of ProcVLM's few-shot adaptation for downstream policy optimization.

\textit{On the real robot}, ProcVLM-guided RFT improves soft success over SFT by 25.0 and 12.5 percentage points at 5k and 10k steps, respectively (Table~\ref{tab:rft_results}).
The larger early-stage gain is expected, as SFT may overfit to noisy demonstration behaviors such as repeated grasp retries (Figure~\ref{fig:jaka_rollout_cases}).
ProcVLM-based advantage conditioning helps reduce the influence of these less useful behaviors during fine-tuning.

\section{Conclusion}

This work presents ProcVLM as a procedure-grounded reward model for robotic learning. By combining subtask-structured supervision with reasoning over remaining steps, ProcVLM learns to estimate progress from procedural context, enabling transfer to new tasks and environments. Results on reward modeling, RoboTwin on-policy RL, and real-robot fine-tuning support its use in downstream policy optimization.

\textbf{Limitations and Future Work.}
ProcVLM learns progress from procedure-defined supervision, so its estimates can be affected by the quality of subtask decomposition and temporal boundary localization~\citep{lea2017temporal,sener2018unsupervised,ji2025robobrain}.
ProcVLM uses a lightweight progress-regression head, leaving calibration and robustness improvements to future work through distributional, stronger regression, or comparison-based objectives \citep{bellemare2017distributional,liang2026robometer,tan2025robodopamine}.
Future work will also broaden real-world evaluation across robot platforms, tasks, and environments, with more comprehensive comparisons against other reward models.

\bibliography{references}
\bibliographystyle{plainnat}
\appendix
\clearpage
\appendix

\startcontents[appendices]

\section*{Appendix Contents}
\printcontents[appendices]{}{1}{}

\section{Annotation Pipeline Details}
\label{app:annotation_details}

This section provides additional details of the annotator models, hierarchical annotation process, pipeline profiling results, and prompt templates used in Section~\ref{sec:pipeline}.

\subsection{Annotator Models}
We use different large VLMs for different annotation stages. For video-level planning and subtask temporal localization, we use Qwen3-VL-235B-A22B-Instruct as the annotator model, since these tasks require long-context video understanding and global procedural reasoning. For frame-level reasoning and grounding annotations, we use InternVL3.5-38B, which provides efficient and reliable single-frame visual reasoning. All generated annotations are post-processed into JSONL files and used to construct procedure-aware VQA tasks for ProcVLM pretraining.

\subsection{Pipeline Execution and Post-processing}
\label{app:pipeline_execution}

The annotation pipeline is implemented as a queue-based asynchronous system to avoid serialized CPU/GPU execution. It consists of four decoupled modules: data reading, CPU-side preprocessing, GPU-side VLM inference, and post-processing. Each module runs independently and communicates through bounded queues, so that slow I/O, image preprocessing, and VLM inference do not block each other. This design overlaps CPU preparation with GPU execution, reduces pipeline bubbles, and keeps the annotator GPUs continuously supplied with ready-to-run batches.

\textbf{Data reader.}
The data reader scans raw robot episodes, organizes them into task-consistent trajectories, and performs sparse video sampling according to the annotation stage. It also prefetches images and metadata asynchronously, including task instructions, camera keys, frame indices, and episode identifiers. This module isolates storage and decoding latency from the rest of the pipeline, preventing GPU workers from waiting on raw data loading.

\textbf{CPU processor.}
The CPU processor converts sampled episodes into VLM-ready inputs. It performs image loading, resizing, multimodal preprocessing, prompt construction, and template formatting for different annotation tasks, including plan generation, subtask temporal localization, frame-level reasoning and target grounding. Because each annotation query often contains multiple images or video frames, we move image loading, multimodal preprocessing, and template formatting to CPU workers ahead of GPU inference, keeping ready-to-run batches available for GPU workers and reducing idle GPU time.

\textbf{GPU worker.}
The GPU worker consumes preprocessed batches and runs large VLM inference. 
Depending on the annotation stage, it performs video-level task planning, subtask temporal localization, or keyframe-level reasoning and grounding. 
We use optimized inference backends such as vLLM and LMDeploy, together with local batched inference APIs, to keep large multimodal requests densely packed on the GPU and maximize annotation throughput.

\textbf{Consumer and post-processing.}
The consumer parses VLM outputs, validates their format, and writes normalized JSONL annotations indexed by dataset name, episode id, frame id, and camera key. It expands video-level subtask segments into frame-wise assignments, propagates keyframe reasoning to neighboring non-keyframes when appropriate, checks temporal consistency for grounding box outputs and removes invalid boxes. This stage converts heterogeneous VLM outputs into a standardized annotation format, allowing different robot datasets to be merged into ProcCorpus and later converted into ProcVQA training samples.

Overall, the pipeline separates I/O-bound, CPU-bound, GPU-bound, and format-normalization workloads into independent stages. This modular design improves annotation throughput, reduces idle time caused by CPU/GPU synchronization, and makes large-scale frame-wise procedural annotation practical over hundreds of thousands of trajectories.

\subsection{Pipeline Profiling}
\begin{figure}
    \centering
    \includegraphics[width=0.95\linewidth]{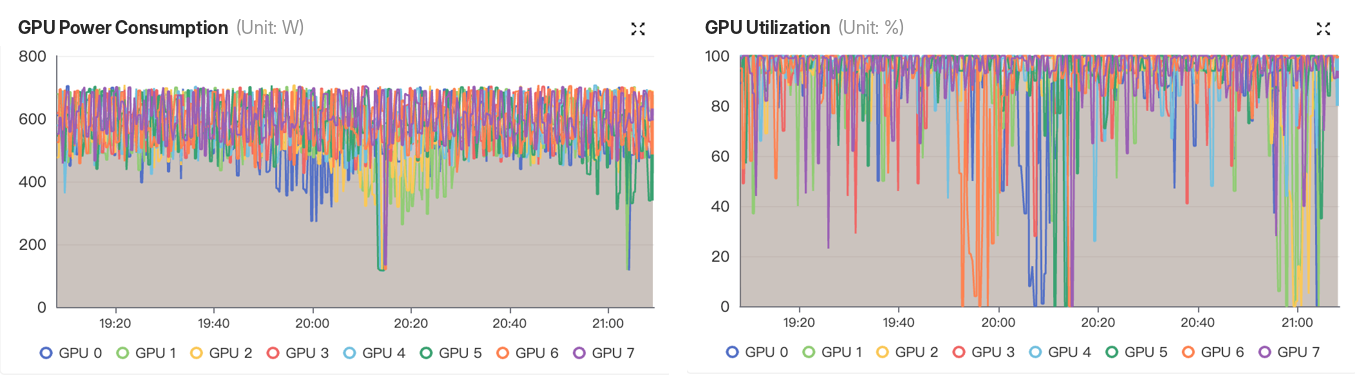}
    \caption{GPU power consumption and utilization over time across multiple GPUs during the annotation pipeline execution. The GPUs maintain consistently high utilization and stable power draw for most of the run, indicating efficient parallel workload scheduling with only brief transient drops.}
    \label{fig:pipeline_profiling}
\end{figure}

In our profiling experiment, the pipeline is deployed on 8 NVIDIA H100 GPUs with 80GB HBM3 memory. Under this setting, it processes about 4M keyframes per day. Figure~\ref{fig:pipeline_profiling} shows the throughput and GPU utilization of the annotation pipeline.

\subsection{Prompt Templates}
We use task-specific prompt templates for plan generation, subtask temporal localization, and frame-level reasoning. The templates are shown below.

\textbf{Plan Generation.}
\begin{small}
\begin{Verbatim}[breaklines=true,breakanywhere=true,breaksymbolleft={}]
I will give you a robot task and a video showing the robot arm performing the task.
You need to analyze actions of the robot arm from the video and decompose the task
into a sequence of detailed sub-tasks. The sequence of sub-tasks should lead to the
completion of the overall task.

###
Grasp [specific object]
e.g. "Grasp the red block"
Explain: Use this pattern when the robot isn't holding anything and needs to pick up
an object, before performing other actions like placing or lifting.
###
Place [specific object] onto / into [specific location]
e.g. "Place the cup onto the table" or "Place the screwdriver into the tool rack."
Explain: Use this pattern when the robot is holding an object and needs to put it
down at a specific location.
###
Push [specific object] [forward / backward / left / right]
e.g., "Push the blue block forward"
Explain: Use this pattern when the robot needs to move an object in a specific
direction by applying force to it, without lifting or grasping it. "Push" and
("Grasp", "Place") are mutually exclusive.
###
Tilt the gripper
e.g. "Tilt the gripper to pour the liquid" or "Tilt the gripper slightly to the left"
Explain: Use this pattern when the robot needs to adjust the angle of its gripper,
either to pour liquid or position the gripper for some specific task.
###
Hang [specific object] on / above [specific location]
e.g. "Hang the coat on the hook" or "Hang the cup above the table"
Explain: Use this pattern when the robot needs to suspend an object from a specific
location, such as hanging a coat on a hook or a cup above a table.
###
Press [specific object]
e.g., "Press the button" or "Press the power switch until it clicks."
###
Open [specific object]
e.g. "Open the door slowly"
###
Close [specific object]
e.g., "Close the lid securely"
###
Rotate [specific object]
e.g., "Rotate the knob clockwise"
###

All sub-tasks must be in exactly one of the patterns above, and should follow the
Explain for each pattern. There is no need to include the robot arm itself as an
object in the sub-tasks.

You should output sub-tasks in a numbered list format, starting from 1. Each line
contains one sub-task with a leading number and a period. No extra text or
explanation.

Task: {task}
Output:
\end{Verbatim}
\end{small}

\textbf{Subtask Segmentation.}
\begin{small}
\begin{Verbatim}[breaklines=true,breakanywhere=true,breaksymbolleft={}]
You will be shown a VIDEO of a robot task and an UNORDERED list of planned sub-tasks.

Task: "{task}"
Planned sub-tasks:
{plans}

OBJECTIVE:
For each planned sub-task, if present, mark the frames where it starts and finishes.
If a sub-task is started but not finished, set complete_frame=null. If not present
at all, set both start_frame=null and complete_frame=null, and notes="not present".
If the video shows any action was interrupted and the overall task was not completed,
set overall_notes="task not completed".

OUTPUT FORMAT:
{
  "task": "<same as input task>",
  "subtasks": [
    {"id":1, "notes":"<<=60 words optional>",
     "start_frame":<int|null>, "complete_frame":<int|null>,
     "name":"<same text from plans>"},
    ...
  ],
  "overall_notes":"<<=30 words optional>"
}

HINTS:
- Find the changes in effector pose and object motion as candidate start/complete frames.
- The start frame can be picked slightly earlier and the complete frame slightly later
  to ensure the action is fully captured.
- If multiple candidates exist, pick the final success. If retries occur, record the
  final success.
- Use the last frame of the video as reference for overall_notes.

Now process the provided video and planned sub-tasks and return the JSON result ONLY.
\end{Verbatim}
\end{small}

\textbf{Frame-level Reasoning.}
We use three reasoning templates depending on the task state: unfinished, finished, and give-up/failure. These templates are intentionally short and task-oriented, encouraging the annotator VLM to focus on task completion and remaining procedural steps rather than open-ended scene description.

\textit{Unfinished state:}
\begin{small}
\begin{Verbatim}[breaklines=true,breakanywhere=true,breaksymbolleft={}]
The image shows a robot performing a task: '{task}', which may be incomplete.
Remaining subtasks: '{rest_sub_task}'.
Explain why it's unfinished and briefly describe the next steps based on image details.

Output (<=150 words, 3 sentences):
<analysis with image details>. This task is not finished <short reason>.
<one-sentence summary of next steps>.

Example:
Task: 'put all the green objects on the pink plate.'
Image: a green apple in robot arm, a green pear on blue plate.
Output:
Image shows a green apple held by the robot and a green pear on the blue plate.
This task is not finished because both green objects are not yet on the pink plate.
The robot should place the green apple on the pink plate, then move the green pear
from the blue plate to the pink plate.
\end{Verbatim}
\end{small}

\textit{Finished state:}
\begin{small}
\begin{Verbatim}[breaklines=true,breakanywhere=true,breaksymbolleft={}]
The image shows a robot performing a task: '{task}', which is finished.
Explain briefly why it's completed based on image details.

Output (<=50 words, 2 sentences):
<analysis with image details>. This task is finished <short reason>.

Example:
Task: 'put all the green objects on the pink plate.'
Image: both green apple and pear on pink plate.
Output:
Image shows a green apple and pear on the pink plate. This task is finished because
all green objects are placed correctly.
\end{Verbatim}
\end{small}

\textit{Give-up or failed state:}
\begin{small}
\begin{Verbatim}[breaklines=true,breakanywhere=true,breaksymbolleft={}]
The image shows a robot performing a task: '{task}', which is not finished.
Explain briefly why it's unfinished based on image details.

Output (<=50 words, 2 sentences):
<analysis with image details>. This task is not finished <short reason>.

Example:
Task: 'put all the green objects on the pink plate.'
Image: a green apple held by the robot, a green pear on blue plate.
Output:
Image shows a green apple in the robot arm and a green pear on the blue plate.
This task is not finished because neither object has been placed on the pink plate.
\end{Verbatim}
\end{small}

{\small
\setlength{\tabcolsep}{3pt}
\setlength{\LTcapwidth}{\textwidth}
\begin{longtable}{@{}p{0.5\textwidth}rrrr@{}}
  \caption{Datasets used for ProcCorpus construction. Annotated frames and annotation rates are computed based on subtask annotations only.}
  \label{tab:dataset_details} \\
  \toprule
  Dataset & Frames & Traj. & \shortstack{Ann.\\Frames} & \shortstack{Ann.\\Rate} \\
  \midrule
  \endfirsthead

  \toprule
  Dataset & Frames & Traj. & \shortstack{Ann.\\Frames} & \shortstack{Ann.\\Rate} \\
  \midrule
  \endhead

  \multicolumn{5}{@{}l}{\textit{Real-robot datasets}} \\
  \midrule
  DROID & 27,630,375 & 95,658 & 26,157,676 & 94.67\% \\
  BridgeData V2 & 2,863,587 & 78,637 & 2,790,172 & 97.44\% \\
  Fractal & 3,786,400 & 87,212 & 3,778,827 & 99.80\% \\
  RH20T & 1,699,138 & 4,433 & 1,621,815 & 95.45\% \\
  Table30 & 5,184,355 & 25,610 & 5,167,461 & 99.67\% \\
  OXE / Austin BUDS & 34,112 & 50 & 34,112 & 100.00\% \\
  OXE / Austin SAILOR & 353,094 & 240 & 353,094 & 100.00\% \\
  OXE / Austin SIRIUS & 279,939 & 559 & 279,939 & 100.00\% \\
  OXE / Berkeley AUTOLab UR5 & 86,887 & 896 & 86,583 & 99.65\% \\
  OXE / Berkeley FANUC Manipulation & 62,613 & 415 & 59,764 & 95.45\% \\
  OXE / CMU Play Fusion & 235,922 & 576 & 235,096 & 99.65\% \\
  OXE / Columbia PushT & 24,802 & 122 & 23,220 & 93.62\% \\
  OXE / DLR EDAN Shared Control & 8,928 & 104 & 8,597 & 96.29\% \\
  OXE / DLR SARA Pour & 12,971 & 100 & 12,771 & 98.46\% \\
  OXE / Dobb-E & 1,139,911 & 5,208 & 1,087,931 & 95.44\% \\
  OXE / FMB & 1,137,340 & 8,611 & 959,915 & 84.40\% \\
  OXE / IAMLab CMU Pickup-Insert & 146,105 & 522 & 145,053 & 99.28\% \\
  OXE / Jaco Play & 70,127 & 976 & 68,872 & 98.21\% \\
  OXE / NYU Door Opening & 17,761 & 435 & 17,670 & 99.49\% \\
  OXE / QUT Dexterous Manipulation & 176,278 & 200 & 175,538 & 99.58\% \\
  OXE / Stanford HYDRA & 358,234 & 570 & 353,326 & 98.63\% \\
  OXE / TOTO & 294,139 & 902 & 294,139 & 100.00\% \\
  OXE / UCSD Kitchen & 3,970 & 150 & 3,491 & 87.93\% \\
  OXE / UT Austin Mutex & 361,883 & 1,500 & 350,737 & 96.92\% \\
  Internal / JAKA Teleop & 23,774 & 100 & 22,271 & 93.68\% \\
  Internal-xArm / Real-world Tabletop & 366,152 & 2,005 & 366,152 & 100.00\% \\
  \midrule
  \textbf{Real-robot total} & \textbf{46,358,797} & \textbf{315,791} & \textbf{44,454,222} & \textbf{95.89\%} \\

  \addlinespace[0.5em]
  \multicolumn{5}{@{}l}{\textit{Simulation datasets}} \\
  \midrule
  GR00T-Teleop-Sim & 5,820,277 & 24,000 & 5,807,497 & 99.78\% \\
  Internal-xArm / Simulation Tabletop & 2,001,729 & 43,720 & 1,998,726 & 99.85\% \\
  LIBERO v2.1 & 273,465 & 1,693 & 273,465 & 100.00\% \\
  RoboTwin 2.0 (\textit{held out from both training stages for OOD evaluation}) & 6,012,086 & 27,200 & 5,997,357 & 99.76\% \\
  \midrule
  \textbf{Simulation total} & \textbf{14,107,557} & \textbf{96,613} & \textbf{14,077,045} & \textbf{99.78\%} \\
  \midrule
  \textbf{Overall total} & \textbf{60,466,354} & \textbf{412,404} & \textbf{58,531,267} & \textbf{96.80\%} \\
  \bottomrule
\end{longtable}
}

\section{Preprocessed Dataset Details}
\label{app:dataset_details}

This section provides the detailed list of real-robot and simulation datasets preprocessed and annotated for ProcCorpus, as described in Section~\ref{sec:data_scaling}. We merge multiple task-level or configuration-level subsets belonging to the same dataset family, and report annotation statistics based only on subtask annotations.

Table~\ref{tab:dataset_details} summarizes the dataset sources and subtask annotation coverage of ProcCorpus. In total, the corpus contains more than 60M raw frames from over 400K real-robot and simulated trajectories, with most datasets achieving high subtask annotation coverage. These statistics show that the annotation pipeline can scale to heterogeneous manipulation data while maintaining dense procedure-level supervision for ProcVLM pretraining.

\begin{figure}[t]
    \centering
    \includegraphics[width=1\linewidth]{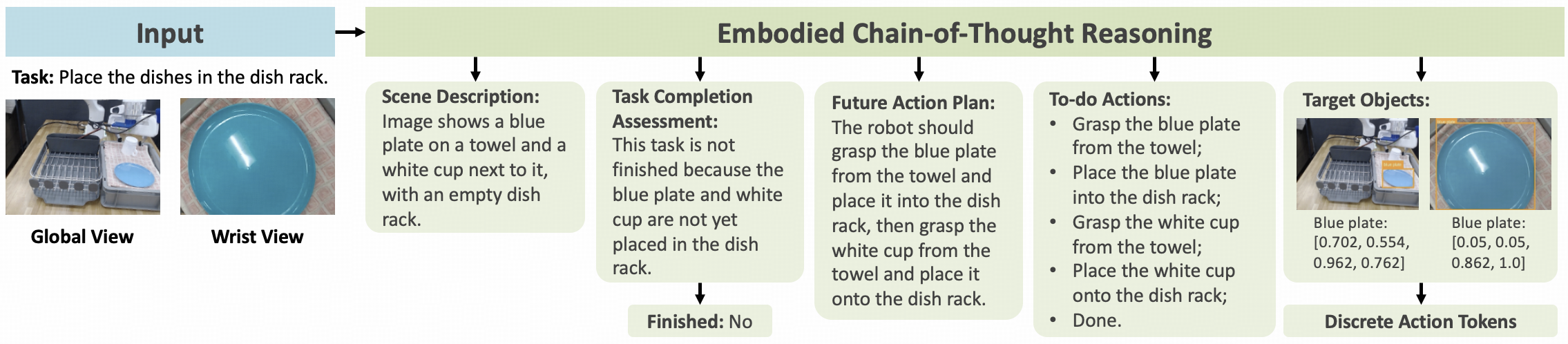}
    \caption{Example of \textit{Embodied Chain-of-Thought (ECoT)} annotation in ProcCorpus. 
    Given multi-view robot observations and a task instruction, ECoT enriches the raw frame with task-centric scene reasoning, completion assessment, future action planning, remaining to-do actions, target-object grounding, and optional discrete action tokens for VLA training.}
    \label{fig:ecot_example}
\end{figure}

As illustrated in Figure~\ref{fig:ecot_example}, each ProcCorpus annotation enriches a raw manipulation frame with task-centric reasoning, completion-state assignment, subtask structure, and target grounding.

\section{ProcVLM Training Details}
\label{app:training_details}

This section provides additional training details for ProcVLM, including the construction of ProcVQA, the model configuration for progress-value prediction, and the two-stage training pipeline, as described in Section~\ref{sec:procvlm_modeling}.

\subsection{ProcVQA Construction}
\label{app:procvqa}

\subsubsection{Construction from Procedural Annotations}

ProcVQA is constructed from the frame-wise annotations in ProcCorpus. 
Each raw trajectory is represented by a task instruction and a robot execution video, where the video is treated as an ordered image sequence. 
The synthesized annotations provide subtask labels, temporal boundaries, frame-level procedural reasoning, and progress estimates.

We convert these annotations into supervised multimodal question-answering samples. 
Each sample consists of a textual query, one or more visual observations sampled from the trajectory, and a target response derived from the corresponding procedural annotations. 
This conversion turns dense frame-wise annotations into a unified VQA-style supervision format for multimodal instruction tuning.

We define an \textbf{atomic action} as an executable subtask that can be described by a \textbf{single explicit verb-level} action. 
A high-level task may admit multiple valid decompositions under this definition, but we regard them as acceptable as long as each atomic action is temporally coherent and verb-maximal, i.e., it covers a complete continuous execution phase whenever possible. 
Under this principle, different decompositions can still provide meaningful low-level and generalizable action descriptions.

For instance, the high-level instruction ``clean the table'' is not directly executable as a single verb-level action. 
When two tissues are on the table, a valid decomposition may include grasping the left tissue, placing it into the trash can, grasping the right tissue, and placing it into the trash can. 
ProcVQA therefore supervises the model to reason over explicit procedural steps rather than only imitate trajectory-level task descriptions.

\subsubsection{Task Templates}

These templates instantiate the pretraining tasks introduced in Section~\ref{sec:procvlm_pretraining}, with five concrete variants denoted as (a.1), (a.2), (b.1), (b.2), and (c). 
They use either video-level interleaved frame-index sequences for action segmentation or recent local observation windows for next-step prediction, remaining-step planning, and progress estimation.

An interleaved frame-index sequence represents a video as an ordered multimodal sequence, where each frame image is followed by its frame identifier, e.g., \texttt{<image><frame\_id: 1><image><frame\_id: 2>}. 
This format preserves the temporal order of sampled frames while remaining compatible with the image-text input interface of the VLM.

\textbf{(a.1) Action segmentation with task instruction.}
Given an interleaved frame-index sequence and its task instruction, the model segments the execution into consecutive atomic actions:
\begin{quote}
\small\ttfamily
Segment the execution of the task "\{task\}" into consecutive atomic actions. Each segment should correspond to a single, explicit verb-level action. Output a JSON list with keys "action\_description", "start\_frame", and "end\_frame".
\end{quote}
The target response is a JSON-style list of action segments, where each segment contains an action description and its start and end frames.

\textbf{(a.2) Task-free action segmentation.}
We also construct a task-free segmentation variant, where the model receives only the visual sequence and predicts the atomic action segments without the original task instruction:
\begin{quote}
\small\ttfamily
Segment the actions shown in the image sequence into consecutive atomic actions, each described by a single explicit verb. Output a JSON list with keys "action\_description", "start\_frame", and "end\_frame".
\end{quote}
This variant encourages the model to infer procedural structure directly from visual observations.

\textbf{(b.1) Immediate next-step prediction.}
Given a recent observation window and the task instruction, the model predicts the immediate next atomic action:
\begin{quote}
\small\ttfamily
Given the recent observation and the task "\{task\}", predict the immediate next atomic action the robot should execute. Use a single explicit verb-level description.
\end{quote}
The target response is a single executable verb-level action.

\textbf{(b.2) Future planning.}
Given a recent observation window and the task instruction, the model predicts the remaining atomic actions required to complete the task:
\begin{quote}
\small\ttfamily
Given the recent observation and the task "\{task\}", list the remaining atomic actions required to complete the task, starting from the next time step. Each action should be a single explicit verb-level step.
\end{quote}
The target response is an ordered list of remaining subtasks.

\textbf{(c) Subtask-structured progress prediction.}
For progress prediction, the model first infers the remaining atomic actions and then estimates the current task completion percentage:
\begin{quote}
\small\ttfamily
Given the recent observation and the task "\{task\}", first infer the remaining atomic actions required to complete the task. Then estimate the current completion percentage and output it as a float wrapped by <progress> tags.
\end{quote}
The response ends with a structured progress tag:
\[
\texttt{<progress>}~p~\texttt{\%</progress>},
\]
where $p$ is a continuous completion percentage. 
This format provides both token-level supervision for procedural reasoning and scalar supervision for the progress value head.

\subsubsection{Training Sample Statistics}

ProcVQA is split into a large-scale pretraining set and a curated refinement set. 
The first-stage set is constructed from the training split of the annotated corpus and provides broad coverage across robots, environments, viewpoints, and task types. 
The second-stage set is built from human-selected trajectories with more accurate subtask alignment and clearer procedural structure.

\begin{table}[htbp]
  \centering
  \small
  \caption{Summary of ProcVQA training data used for the two-stage ProcVLM training pipeline.}
  \label{tab:procvqa_training_stats}
  \begin{tabular}{lcc}
    \toprule
    Statistic & Stage 1 Pretraining & Stage 2 Refinement \\
    \midrule
    Data source & ProcVQA training split & Human-selected subset \\
    Token scale & $\sim$20B & 2.8B \\
    Trajectory scale & $\sim$400K & $\sim$15K \\
    Main purpose & Broad procedural pretraining & High-quality refinement \\
    \bottomrule
  \end{tabular}
\end{table}

To further characterize the procedural structure of the curated refinement set, we report the distribution of subtask counts per trajectory in Table~\ref{tab:stage2_subtask_count}. 
The Stage 2 set contains 13,688 trajectories, with an average of 2.87 subtasks per trajectory and a median of 3. 
This indicates that the curated set contains substantial multi-step procedural structure rather than only single-action executions, supporting its use for refinement-stage training on subtask alignment and progress reasoning.

\begin{table}[htbp]
  \centering
  \small
  \caption{Distribution of subtask counts per trajectory in the Stage 2 curated refinement set.}
  \label{tab:stage2_subtask_count}
  \setlength{\tabcolsep}{4pt}

  \begin{tabular}{lrrrrrrrrrrrrr}
    \toprule
    \# Subtasks 
      & 1 & 2 & 3 & 4 & 5 & 6 & 7 & 8 & 9 & 10 & 11 & 12 & $\ge14$ \\
    \midrule
    \# Trajectories 
      & 1455 & 5303 & 4187 & 1523 & 210 & 479 & 123 & 197 & 19 & 153 & 2 & 15 & 22 \\
    \bottomrule
  \end{tabular}

  \vspace{0.6em}

  \begin{tabular}{lc}
    \toprule
    Summary statistic & Value \\
    \midrule
    Total trajectories & 13,688 \\
    Mean \# subtasks & 2.87 \\
    Median \# subtasks & 3 \\
    Minimum \# subtasks & 1 \\
    Maximum \# subtasks & 24 \\
    \bottomrule
  \end{tabular}
\end{table}

\subsection{ProcVLM Configuration}
\label{app:procvlm_setting}

\subsubsection{Backbone and Output Branches}

ProcVLM is initialized from Qwen3-VL-2B-Instruct. 
The backbone takes visual observations and task instructions as multimodal input and generates textual responses through the standard autoregressive language modeling head. 
This language branch is used for all ProcVQA task families.

The value head is implemented as a three-layer MLP with hidden dimensions $d_h \rightarrow 4d_h \rightarrow d_h \rightarrow 1$, where $d_h$ denotes the hidden size of the VLM backbone.

\subsubsection{Progress-Value Gating}

The progress value head is controlled by a semantic gating mechanism. 
During training, this branch is activated when the ground-truth response contains a \texttt{<progress>} tag. 
For samples without progress supervision, only the language modeling objective is applied. 
During inference, the value branch can be activated on demand for responses containing a \texttt{<progress>} tag. This gated design separates general procedure-aware text generation from continuous value prediction. 
It allows ProcVLM to learn from all task families through language modeling, while using progress-prediction samples to additionally supervise the value head.

\subsubsection{Value-Head Pooling}

Let $H = \{h_i\}_{i=1}^{L}$ denote the sequence hidden states produced by the VLM backbone. 
For progress-prediction samples, the value head applies attention pooling over the valid sequence positions:
\[
\alpha_i
=
\frac{
\exp(s(\tilde{h}_i)) \cdot m_i
}{
\sum_{j=1}^{L}
\exp(s(\tilde{h}_j)) \cdot m_j
},
\]
\[
h_{\mathrm{pool}}
=
\sum_{i=1}^{L} \alpha_i \tilde{h}_i,
\qquad
\hat{p}
=
f_{\mathrm{value}}(h_{\mathrm{pool}}),
\]
where $s(\cdot)$ is the attention scoring function, $f_{\mathrm{value}}$ is the regression head, and $m_i$ denotes the attention mask that excludes padding tokens. 
During training, $\tilde{h}_i$ is obtained by applying feature-level dropout to the tail hidden states that contain the progress answer, while keeping other hidden states unchanged. 
This corruption reduces direct access to the ground-truth progress tokens during teacher forcing and encourages the value head to infer progress from the visual observation, task instruction, and prior procedural reasoning.

\subsection{Training Pipeline and Implementation Details}
\label{app:procvlm_training}

\subsubsection{Training Objectives}
\textbf{Language Modeling Loss.}
Given the ProcVQA templates above, all task families are trained with the standard autoregressive language modeling objective over the supervised assistant response. 
The input prompt, including task instructions and visual observations, is used as conditioning context, while the loss is applied only to the target response tokens. 
Formally, given a multimodal input $x$ and a target response $y = \{y_t\}_{t=1}^{T}$, we compute
\[
\mathcal{L}_{\mathrm{LM}}
=
-\frac{1}{|\mathcal{M}_{\mathrm{text}}|}
\sum_{t \in \mathcal{M}_{\mathrm{text}}}
\log P_{\theta}(y_t \mid y_{<t}, x),
\]
where $\mathcal{M}_{\mathrm{text}}$ denotes the set of target response tokens used for language modeling supervision. 
Prompt tokens and padding tokens are masked out in the label tensor and are therefore excluded from the loss.

Progress-prediction samples are also included in this language modeling objective. 
In these samples, the reasoning context and the formatted progress answer are learned as text, while the scalar value inside the \texttt{<progress>} tag is additionally used as supervision for the value head.

\textbf{Progress Regression Loss.}
For progress-prediction samples, we parse the scalar completion value from the structured \texttt{<progress>} tag and use it as the ground-truth progress label. 
Although the textual response expresses progress as a percentage, the regression loss is computed on a normalized scale. 
Let $p$ denote the parsed completion percentage and $\bar{p}=p/100$ denote its normalized value. 
The value head predicts a normalized progress score $\hat{p}\in[0,1]$, and the value loss is defined as
\[
\mathcal{L}_{\mathrm{value}}
=
\ell_{\mathrm{reg}}(\hat{p}, \bar{p}),
\]
where $\ell_{\mathrm{reg}}$ is the regression loss used for continuous progress prediction.

\textbf{Joint Objective.}
The overall objective is
\[
\mathcal{L}
=
\mathcal{L}_{\mathrm{LM}}
+
\lambda \cdot \mathbb{I}_{\mathrm{prog}} \mathcal{L}_{\mathrm{value}},
\]
where $\mathbb{I}_{\mathrm{prog}}$ indicates whether the sample contains progress supervision and $\lambda$ controls the contribution of the value regression loss. 
In minibatches containing both progress and non-progress samples, the value loss is averaged only over samples with valid progress labels.

\subsubsection{Tail Dropout for Leakage Mitigation}

During teacher forcing, the ground-truth progress value appears near the end of the textual target. 
If the value head directly relies on the hidden states of the \texttt{<progress>} answer tokens, it may recover the target value from the answer span rather than infer task progress from the visual observation and procedural context. 
To reduce this leakage, we apply a soft tail-masking strategy before value-head pooling.

Specifically, for each training sequence, we identify the last valid token according to the attention mask and apply feature-level dropout to the final $N_{\mathrm{tail}}$ hidden states:
\[
\tilde{h}_i =
\begin{cases}
\mathrm{Dropout}(h_i; p_{\mathrm{tail}}), & i \in \mathcal{T}_{\mathrm{tail}}, \\
h_i, & \text{otherwise},
\end{cases}
\]
where $\mathcal{T}_{\mathrm{tail}}$ denotes the tail token region containing the progress answer. 
In our implementation, we set $N_{\mathrm{tail}}=14$, which covers the numerical progress answer, the surrounding \texttt{<progress>} tags, and the sequence ending tokens in the default response format. 
We use $p_{\mathrm{tail}}=0.5$ for feature-level dropout during training.

The corrupted hidden states $\tilde{H}=\{\tilde{h}_i\}_{i=1}^{L}$ are then passed to the attention pooler together with the standard attention mask. 
Importantly, the tail tokens are not removed from pooling by setting their attention mask to zero. 
Instead, their hidden features are partially corrupted. 
This soft masking weakens direct access to the ground-truth progress tokens while keeping the tail positions in the computation graph, allowing gradients to still flow through the complete response sequence.

At inference time, no tail dropout is applied. 
When the value branch is enabled, the model runs a forward pass over the generated sequence and predicts the progress value from the resulting hidden states.

\subsubsection{Two-stage Training Pipeline}

\begin{figure}
    \centering
    \includegraphics[width=1\linewidth]{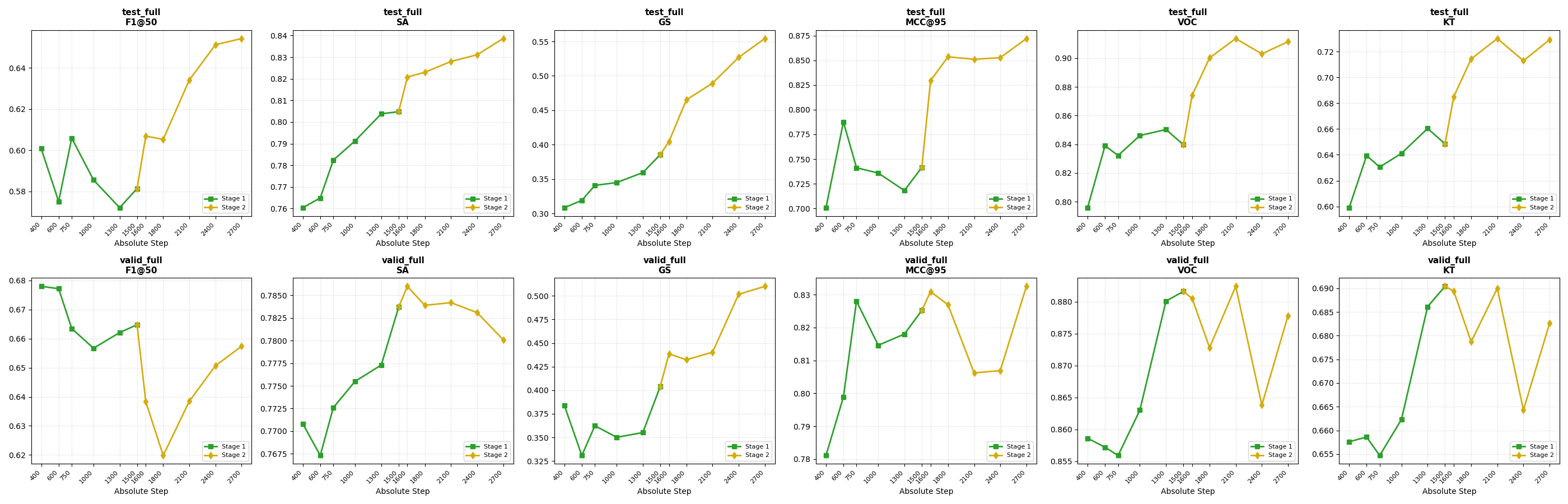}
    \caption{Training curves of ProcVLM under the two-stage training pipeline. Green denotes Stage 1 pretraining on the ProcVQA training split, and yellow denotes Stage 2 refinement on the curated subset. Metrics are evaluated on the in-distribution \texttt{test\_full} split and the out-of-distribution \texttt{valid\_full} split. Stage 1 mainly establishes generalizable procedural representations from large-scale diverse supervision, while Stage 2 further improves prediction precision with higher-quality refinement data.}
    \label{fig:two_stage_training_curves}
\end{figure}

As described in Section~\ref{sec:training_objectives}, ProcVLM is trained in two stages: large-scale pretraining on the ProcVQA training split, followed by refinement on a curated subset constructed from human-selected trajectories. 
We provide the training curves in Figure~\ref{fig:two_stage_training_curves} to analyze the distinct roles of the two stages.

Figure~\ref{fig:two_stage_training_curves} reports representative metrics throughout training on both the in-distribution \texttt{test\_full} split and the out-of-distribution \texttt{valid\_full} split. 
F1@50 directly measures action segmentation quality by matching predicted and ground-truth subtask intervals. 
SA evaluates the semantic accuracy of predicted subtask plans. 
GS measures progress regression accuracy. 
MCC@95 evaluates success-state prediction by converting progress into a binary decision, where $p \geq 95$ and $\hat{p} \geq 95$ indicate ground-truth and predicted success, respectively. 
VOC and KT both measure intra-trajectory ordering ability, i.e., whether predicted progress values preserve the temporal order of states within a trajectory.

The Stage 1 curves show that large-scale pretraining is crucial for generalization. 
Training on the ProcVQA training split consistently improves planning, progress estimation, success prediction, and trajectory-ordering metrics on both in-distribution and out-of-distribution splits, indicating that diverse procedure-aware supervision helps learn transferable task-progress representations.

Stage 2 further refines the pretrained model on the curated subset. 
After switching to higher-quality data, the model gains precision on metrics requiring accurate subtask boundaries, calibrated progress values, and reliable temporal ordering, with stronger improvements on the in-distribution split and smaller gains or fluctuations on the out-of-distribution split. 
Together, these trends support the two-stage design: Stage 1 builds generalizable procedural representations, while Stage 2 sharpens subtask and progress prediction.

\textbf{Training Hyperparameters.} 
The main training hyperparameters are summarized in Table~\ref{tab:training_hyperparams}. 
Both stages use the same joint objective described above and train the VLM backbone together with the progress value head. 
They differ mainly in data distribution, optimization schedule, and value-head regularization: Stage 1 uses a larger learning rate and lower value-head regularization for broad pretraining, while Stage 2 uses a smaller learning rate and stronger refinement regularization on the curated subset.

\begin{table}[htbp]
  \centering
  \small
  \caption{Training hyperparameters for the two-stage ProcVLM training pipeline.}
  \label{tab:training_hyperparams}
  \setlength{\tabcolsep}{4pt}
  \begin{tabular}{p{0.32\linewidth} p{0.29\linewidth} p{0.29\linewidth}}
    \toprule
    Hyperparameter & Stage 1 Pretraining & Stage 2 Refinement \\
    \midrule
    Training data 
      & \texttt{train\_s1\_8k} 
      & \texttt{train\_s2\_8k} \\
    Token scale 
      & $\sim$20B 
      & 2.8B \\
    Backbone initialization 
      & Qwen3-VL-2B-Instruct 
      & Stage 1 checkpoint \\
    Trainable modules 
      & VLM backbone, vision tower, multimodal projector, value head 
      & VLM backbone, vision tower, multimodal projector, value head \\
    LoRA 
      & Disabled 
      & Disabled \\
    Precision 
      & bf16 
      & bf16 \\
    Context length 
      & 8192 
      & 8192 \\
    Data packing 
      & Enabled 
      & Enabled \\
    Per-device batch size 
      & 1 
      & 1 \\
    Gradient accumulation 
      & 64 
      & 64 \\
    Effective batch size 
      & $64 \times N_{\mathrm{GPU}}$ 
      & $64 \times N_{\mathrm{GPU}}$ \\
    Base learning rate 
      & $1\times10^{-5}$ 
      & $5\times10^{-6}$ \\
    Multimodal projector learning rate 
      & $1\times10^{-5}$ 
      & $4\times10^{-6}$ \\
    Vision tower learning rate 
      & $5\times10^{-6}$ 
      & $2\times10^{-6}$ \\
    Optimizer 
      & AdamW 
      & AdamW \\
    Learning-rate schedule 
      & Cosine 
      & Cosine \\
    Warmup ratio 
      & 0.03 
      & 0.05 \\
    Weight decay 
      & 0.01 
      & 0.01 \\
    Training length 
      & $\sim$1.1--1.2 epochs 
      & $\sim$3--4 epochs \\
    Image pixel range 
      & $32^2$--$512^2$ 
      & $32^2$--$512^2$ \\
    Video sampling 
      & 2 FPS, up to 512 frames 
      & 2 FPS, up to 512 frames \\
    Regression loss $\ell_{\mathrm{reg}}$ 
      & $L_1$ 
      & $L_1$ \\
    Value-loss weight $\lambda$ 
      & 0.2 
      & 0.3 \\
    Value-head dropout 
      & 0 
      & 0.05 \\
    Value noise std. 
      & 0.03 
      & 0.01 \\
    Distributed training 
      & DeepSpeed ZeRO-2 
      & DeepSpeed ZeRO-2 with offload \\
    Gradient checkpointing 
      & Enabled 
      & Enabled \\
    \bottomrule
  \end{tabular}
\end{table}

\section{ProcVQA Evaluation Details}
\label{app:vlm_details}

This section describes the ProcVQA benchmark and evaluation metrics used in Section~\ref{sec:vlm_exp}, together with the reward-model evaluation protocol for the ProcVQA comparison in Section~\ref{sec:progress_reward_eval}.

\subsection{Benchmark and Splits}
ProcVQA is a human-selected ECoT-based VQA benchmark for embodied procedural understanding. It covers three manipulation-understanding tasks: action segmentation, future planning, and task progress estimation. To evaluate both in-domain performance and cross-domain generalization, we divide ProcVQA into an in-distribution (ID) split and an out-of-distribution (OOD) split. The ID split is derived from training-domain datasets, including DROID, Bridge, Table30, and selected OXE subsets, while the OOD split is constructed from unseen RoboTwin tasks.

\subsection{Metrics}\label{app:procvqa_metrics}
\textbf{Boundary F1 score (BF1).}
BF1 is calculated as follows. Given predicted and ground-truth action segments, we first extract segment boundaries and remove duplicated boundary positions. A predicted boundary is matched to a ground-truth boundary if their temporal distance is within 5\% of the sequence length. Each boundary can be matched at most once. BF1@5 is then computed as the boundary-level F1 score from matched and unmatched boundaries. We use BF1@5 as the primary segmentation metric because it penalizes both missing and redundant boundaries.

\textbf{Matched boundary localization error (mMAE).}
mMAE measures the temporal localization error of matched boundaries. It is computed as the mean absolute distance between each matched predicted boundary and its corresponding ground-truth boundary. Since mMAE is calculated only over matched boundaries, it does not penalize false positives or false negatives. We therefore report it as an auxiliary metric rather than the primary action segmentation metric.

\textbf{Future planning success (Success).}
Success measures whether the model-generated future task plan can be successfully executed and satisfy the given instruction. We compute this metric through anonymized and randomly ordered human evaluation, where annotators compare model outputs without knowing the model identity.

\textbf{Value-Order Correlation (VOC).}
VOC evaluates within-trajectory progress ordering. We compute Spearman's rank correlation between predicted progress values and the temporal order of the corresponding observations. It does not measure numerical agreement with the procedure-defined progress targets.

\textbf{Effective Progress Resolution (EPR).}
EPR measures the density of continuous progress regression. It discourages models from obtaining high VOC by predicting only a few discrete progress anchors, such as $0.25$, $0.5$, $0.75$, and $1.0$. For predicted progress values $\hat{p}$, EPR is defined as
\begin{equation}
\label{eq:effective_progress_resolution}
\mathrm{EPR}_{\tau}(\hat{p}) =
-\log_2 \min \left\{ \Delta \in \{1/k \mid k \in \mathbb{N}_{+}\}
\;\middle|\;
\Delta \cdot |\mathcal{B}_{\Delta}(\hat{p})| \ge \tau
\right\},
\end{equation}
where $\mathcal{B}_{\Delta}(\hat{p})$ denotes the set of occupied quantization bins after quantizing $\hat{p}$ with bin width $\Delta$. In our experiments, we report EPR@50 by setting $\tau=0.5$.

\subsection{Reward-Model Evaluation}
\label{app:baseline_details}

This subsection details the zero-shot reward-model comparison on the ProcVQA-OOD progress-estimation subset in Section~\ref{sec:reward_model_exp}. We use the benchmark split and metrics described above, and specify the input construction and progress-score extraction for each baseline below.

\textbf{Evaluation setting.}
We compare ProcVLM with Robometer and RoboDopamine using short local-window queries, typically containing 1--16 consecutive frames within about one second. Queries from the same trajectory are evaluated independently without chronological ordering. We report VOC and EPR@50 as defined above.

\textbf{RoboDopamine.}
RoboDopamine trains a step-aware vision-language reward model to estimate fine-grained manipulation progress from multi-view state observations by predicting relative progress hops and fusing them into dense rewards. For each local-window query, we retain the temporal order of its frames and prepend a blank image as a neutral contrastive start anchor. This construction provides the contrastive input required by RoboDopamine without introducing observations from other query windows. We use the 4B variant of Robo-Dopamine-GRM-2.0 for this comparison.

\textbf{Robometer.}
Robometer trains a Qwen3VL-based video-language reward model with both frame-level progress/success supervision and trajectory-comparison preference supervision, enabling dense reward prediction from robot videos and language instructions. In our evaluation, Robometer receives the instantaneous input window from each VQA query, and we use its last-frame prediction as the final progress estimate.

\section{RoboFAC Evaluation Details}
\label{app:robofac_details}

This section provides additional details of the RoboFAC evaluation protocol and metric definitions used in Section~\ref{sec:robofac_exp}.

\subsection{Implementation Details}

\textbf{Benchmark.}
RoboFAC is a failure-centric robotic manipulation VQA benchmark for robotic failure analysis and correction. It contains diverse successful and failed manipulation executions, with structured annotations for task understanding, failure analysis, and correction. RoboFAC consists of both real-robot and simulated data, and we use only the real-robot subset for evaluation. We use RoboFAC to evaluate whether ProcVLM can generalize to unseen tasks with only a small number of demonstrations.

\textbf{One-shot split construction.}
We first split all trajectories into success and failure sets. We then construct task-level one-shot training sets separately from these two sets. For the success set, we select one trajectory for each task as the one-shot successful demonstration. For the failure set, we select one trajectory for each task--failure-type pair, so that different failure modes are covered with minimal supervision. All remaining trajectories are used for testing.

\textbf{Viewpoint coverage.}
The one-shot demonstrations are selected at the trajectory level and do not explicitly cover all camera viewpoints. As a result, the test set may contain observations from viewpoints that are absent from the one-shot demonstrations. This design makes the evaluation more challenging and better reflects practical reward-model deployment, where a small number of demonstrations cannot exhaustively cover all visual variations.

\textbf{Success detection.}
For binary success detection, each model performs inference on the full test set using the visual observation from the last temporal window of each trajectory. Robometer predicts completion with its success head, whose probability is discretized into a binary success label. For RoboDopamine and ProcVLM, we use the predicted progress value and apply a fixed success threshold to obtain the binary completion label.

\textbf{Full-trajectory progress evaluation.}
For full-trajectory progress evaluation, running all test trajectories is computationally expensive. We therefore sample 100 test trajectories in total, evenly from the success and failure sets, and evaluate progress prediction along the sampled trajectories.

\subsection{Metrics}

\textbf{VOC on successful trajectories.}
VOC$_{\mathrm{succ}}$ is computed only on successful trajectories and follows the same definition as VOC in ProcVQA. It measures whether predicted progress values preserve the correct temporal order along a successful execution.

\textbf{MAE on failed trajectories.}
MAE$_{\mathrm{fail}}$ is computed only on failed trajectories. For each failed trajectory, we identify the earliest frame that reaches the maximum predicted progress value:
\[
t^{*} = \min \arg\max_t \hat{p}_t,
\]
where $\hat{p}_t$ denotes the predicted progress value at frame $t$. We treat $t^{*}$ as the predicted turning point, corresponding to the estimated failure onset. Given the human-annotated cutoff index $t_{\mathrm{cut}}$, MAE$_{\mathrm{fail}}$ is computed as
\begin{equation}
\label{eq:mae_fail}
\mathrm{MAE}_{\mathrm{fail}} =
\frac{1}{N_{\mathrm{fail}}}
\sum_{i=1}^{N_{\mathrm{fail}}}
\frac{\left|t^{*(i)}-t_{\mathrm{cut}}^{(i)}\right|}{T_i-1}.
\end{equation}
Here, $T_i \geq 2$ is the number of evaluated frames in trajectory $i$,
and both cutoff indices are zero-based.
The error is normalized within each trajectory before averaging across failed trajectories.

\textbf{MCC for success detection.}
We use the Matthews correlation coefficient (MCC) to evaluate binary success detection.
Predicted progress scores or success probabilities are converted into completion labels using a fixed threshold, and MCC is then computed against the ground-truth success/failure labels.

\textbf{Latency.}
Latency is measured in seconds during full-trajectory progress evaluation and reports the average inference time required to process one trajectory. We measure latency on NVIDIA RTX A6000 GPUs, using single-GPU deployment for all models except Qwen3.5-27B, which is deployed with 4-way tensor parallelism.

\section{Reward-guided Policy Optimization Details}
\label{app:policy_optimization_details}

This section provides additional details on the reward-guided policy optimization experiments in Section~\ref{sec:rft_exp}.

\subsection{On-Policy Reinforcement Learning on RoboTwin}
\label{app:robotwin_sim_setup}
\textbf{Tasks and shared RL setup.}
We optimize OpenVLA-OFT with GRPO in RLinf on Place Container Plate and Move Can Pot.
Within each task, all methods share the SFT initialization and policy training/evaluation settings.
Each update uses 480 trajectories (48 environments $\times$ 10 rollout epochs; 60 groups of eight), with rewards computed after each 25-step action chunk.

\textbf{Reward-model adaptation and inference.}
ProcVLM and RoboDopamine use two stage-annotated demonstrations per task; Robometer uses 50 without manual annotations.
For ProcVLM and Robometer, \texttt{w8} denotes an ordered window of up to eight front-camera frames.
\begin{itemize}[leftmargin=1.2em, labelsep=0.4em, itemsep=0.3em, topsep=0.3em]
    \item \textbf{ProcVLM (procedure-grounded, w8).}
    We LoRA-tune on 335 procedure-grounded examples for Move Can Pot and 344 for Place Container Plate; progress is parsed from the model's generated output.
    \item \textbf{Robometer (linear, w8).}
    We LoRA-tune Robometer-4B on 45 demonstrations, holding out five for validation, with the official rank 32 and $\alpha=64$ configuration. Progress labels discretize normalized trajectory time into ten bins.
    \item \textbf{RoboDopamine (Forward + Backward + Incremental, multi-view).}
    We fully fine-tune on 2,000 comparison samples per task using synchronized front and left/right wrist images; inference fuses all three modes across these views.
    \item \textbf{RoboDopamine (Forward + Backward, single-view).}
    On Move Can Pot, we also evaluate forward--backward fusion using only the front camera.
\end{itemize}

\textbf{Reward construction and optimization.}
Progress predictions are normalized to $[0,1]$; consecutive differences are clipped to $[-0.2,0.2]$.
The progress reward has weight 1.0 and the environment success reward 0.1, emphasizing the learned signal.
Following RLinf's official configuration, the policy uses LoRA rank 32, with global batch size 480, weight decay 0.01, and gradient clipping at 1.0.
GRPO uses lower/upper clipping parameters 0.2/0.28, sampling temperature 1.6, and zero KL and entropy coefficients.
The learning rate, discount factor, and episode horizon are $(10^{-5},0.99,150)$ for Place Container Plate and $(2\times10^{-4},1.0,200)$ for Move Can Pot.

\textbf{Evaluation and active-time comparison.}
Every five updates, we evaluate 48 trajectories and report episodic success (\texttt{eval/success\_once}).
We compare success against active time on both tasks, using the first recorded evaluation as the time origin.
Active time excludes downtime associated with policy restarts and includes evaluation and checkpointing overhead, with all models following the same evaluation and checkpointing protocol.

\textbf{Curve presentation.}
In Figure~\ref{fig:robotwin_onpolicy}, dashed lines show raw evaluation success rates; solid lines show rates smoothed by a three-point median filter followed by a weighted moving average, assigning weights of 0.25, 0.5, and 0.25 to the previous, current, and next points, respectively.

\begin{figure}[t]
    \centering

    \begin{minipage}{\linewidth}
        \centering
        \includegraphics[width=\linewidth]{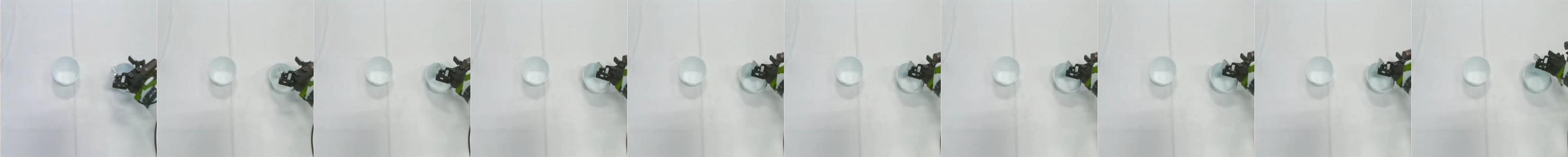}
        \caption*{\footnotesize SFT, 5k training steps, grasp failed.}
    \end{minipage}

    \begin{minipage}{\linewidth}
        \centering
        \includegraphics[width=\linewidth]{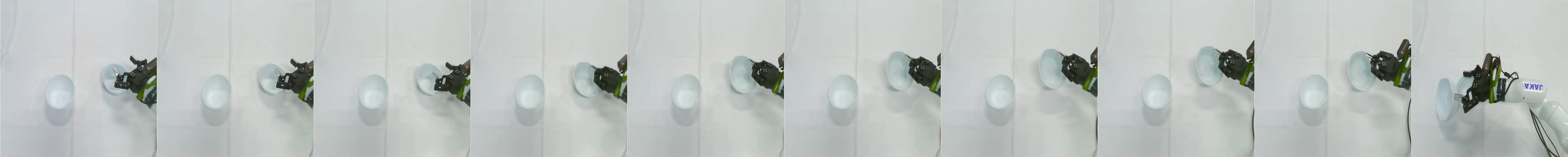}
        \caption*{\footnotesize SFT, 5k training steps, partial success.}
    \end{minipage}

    \begin{minipage}{\linewidth}
        \centering
        \includegraphics[width=\linewidth]{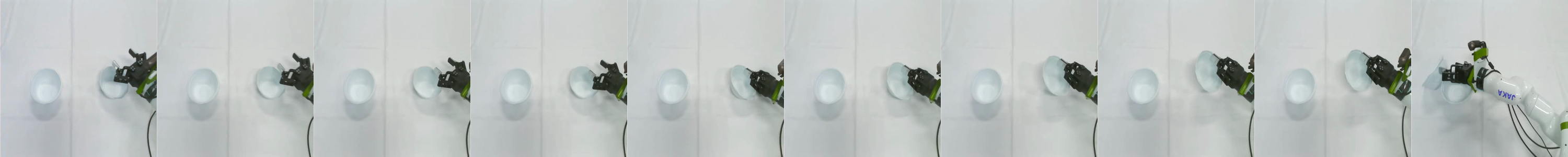}
        \caption*{\footnotesize RFT, 5k training steps, partial success.}
    \end{minipage}

    \begin{minipage}{\linewidth}
        \centering
        \includegraphics[width=\linewidth]{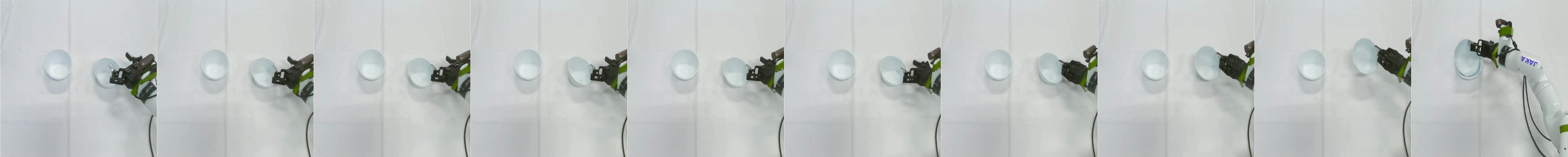}
        \caption*{\footnotesize RFT, 5k training steps, full success.}
    \end{minipage}

    \begin{minipage}{\linewidth}
        \centering
        \includegraphics[width=\linewidth]{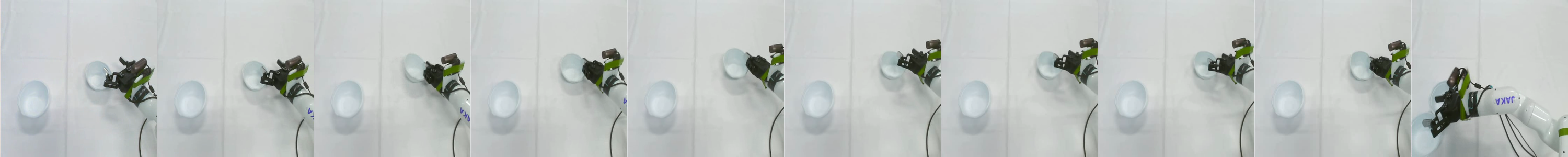}
        \caption*{\footnotesize SFT, 10k training steps, partial success.}
    \end{minipage}

    \begin{minipage}{\linewidth}
        \centering
        \includegraphics[width=\linewidth]{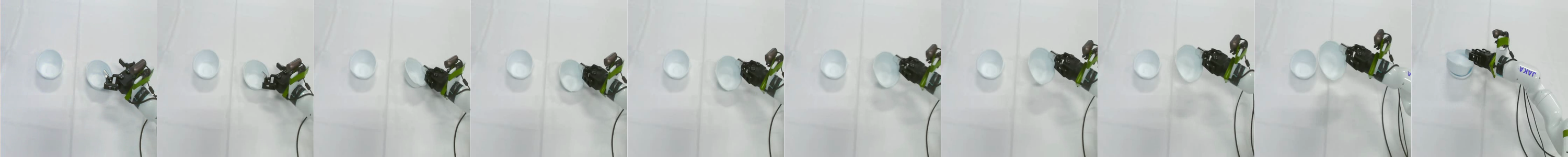}
        \caption*{\footnotesize SFT, 10k training steps, full success.}
    \end{minipage}

    \begin{minipage}{\linewidth}
        \centering
        \includegraphics[width=\linewidth]{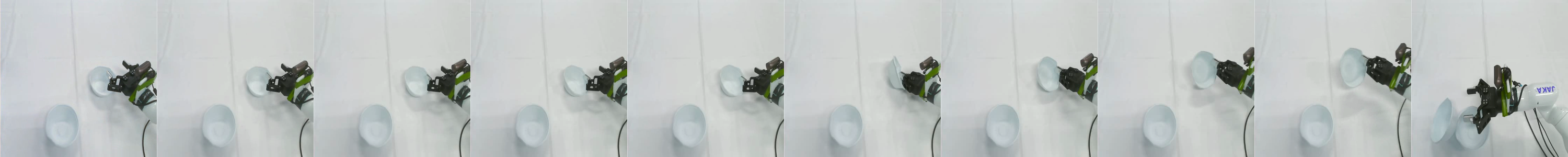}
        \caption*{\footnotesize RFT, 10k training steps, partial success.}
    \end{minipage}

    \begin{minipage}{\linewidth}
        \centering
        \includegraphics[width=\linewidth]{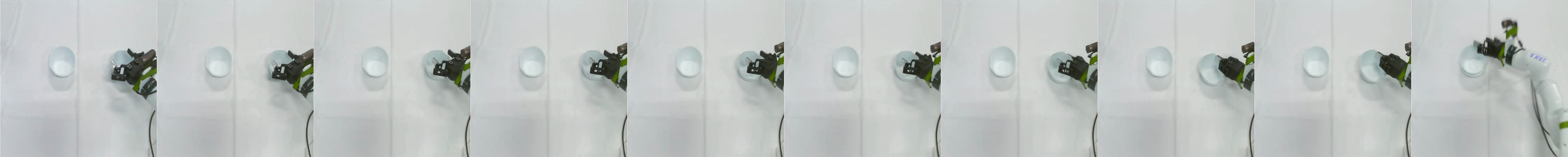}
        \caption*{\footnotesize RFT, 10k training steps, full success.}
    \end{minipage}
    \caption{Representative real-robot rollout records on the JAKA tabletop stack-bowls task. We show examples from SFT and RFT policies at both 5k and 10k training steps. Each method is evaluated with 12 rollouts under the two-bowl setting, and the final score is computed from the nesting quality of the placed bowl.}
    \label{fig:jaka_rollout_cases}
\end{figure}

\subsection{Real-Robot Reward Fine-tuning}
\label{app:jaka_real_robot}

\textbf{Task and data setup.}
We conduct real-robot experiments on a tabletop stack-bowls task with a single-arm JAKA robot.
The task requires the robot to pick up a bowl on the table and place it into a target bowl, forming a properly nested bowl stack.
The real-robot training set contains 50 human teleoperated demonstration trajectories, approximately evenly distributed across scenes with two, three, and four bowls.
We train both SFT and RFT policies for 10k steps with a batch size of 64, and visualize representative rollouts at both 5k and 10k training steps.
Due to the higher difficulty of the three-bowl and four-bowl settings and our limited real-robot evaluation budget, preliminary runs showed very low success rates on these settings.
Therefore, we restrict the real-robot evaluation to the two-bowl setting, where each rollout requires one bowl to be placed into the target bowl.

\textbf{Reward-guided fine-tuning.}
Both SFT and RFT start from the same $\pi_{0.5}$ policy initialization and use the same demonstration data.
We use SJTU Evo-RL, an offline RL framework supporting value inference and advantage-conditioned policy training.
SFT follows standard supervised fine-tuning. For RFT, ProcVLM assigns progress scores as dense rewards to the training trajectories, and Evo-RL estimates advantages over a 50-step horizon.
The top 30\% of samples by advantage receive positive labels and the remaining samples receive negative labels; these labels serve as auxiliary conditions during fine-tuning.
The teleoperated demonstrations contain noisy local behaviors, including repeated grasp retries. Advantage conditioning helps distinguish higher-advantage segments from less useful behaviors, offering a possible explanation for the larger RFT gain at 5k steps.

\textbf{Evaluation protocol.}
For each method, we conduct 12 real-robot rollouts under the two-bowl setting.
Success is measured by the degree of proper nesting in the final bowl configuration.
For each rollout, we assign a nesting score $s \in [0,1]$ according to the following rubric:
\begin{itemize}[leftmargin=1.2em, labelsep=0.4em, itemsep=0.2em, topsep=0.2em]
    \item $s=0.0$ \textbf{(failure)}: the stacking action is not completed within 30 seconds, such as when the bowl is not grasped or the robot does not attempt insertion.
    \item $s=0.5$ \textbf{(partial success)}: the robot attempts to insert the bowl, but the placed bowl slips out of the target bowl or is placed immediately next to the target bowl.
    \item $s=1.0$ \textbf{(full success)}: the robot successfully places the bowl inside the target bowl.
\end{itemize}
The final score of each method is computed as the average nesting score over 12 rollouts.
We report this value as the soft success rate, where partial placements receive half credit.

\textbf{Rollout visualization.}
Figure~\ref{fig:jaka_rollout_cases} shows representative real-robot rollout records under the two-bowl evaluation setting.
These examples include different execution outcomes and illustrate how the final nesting score is assigned based on the resulting bowl configuration.

\begin{table}[H]
  \centering
  \small
  \setlength{\tabcolsep}{6pt}
  \renewcommand{\arraystretch}{1.12}
  \caption{Real-robot experimental setup on the JAKA stack-bowls task.}
  \label{tab:jaka_real_robot_setup}
  \begin{tabular}{ll}
    \toprule
    Item & Description \\
    \midrule
    Robot platform & JAKA robot, single arm \\
    Scene & Tabletop stack-bowls task \\
    Training data & 50 human teleoperated demonstrations \\
    Demonstration settings & Two-bowl, three-bowl, and four-bowl scenes \\
    Evaluation setting & Two-bowl scene only \\
    Rollouts per method & 12 rollouts \\
    Metric & Soft success rate \\
    \bottomrule
  \end{tabular}
\end{table}

\section{Additional Analyses}
\label{app:additional_analyses}

\subsection{Annotation Quality Audit}
\label{app:annotation_quality}

We manually audit the automatically generated subtask annotations for 92,465 trajectories under a strict trajectory-level criterion.
A trajectory qualifies only if all subtask descriptions are correct and all temporal boundaries align with the video within a tolerance of one second.
Any incorrect subtask description or boundary deviation exceeding one second causes the entire trajectory to be rejected.
The resulting pass rate therefore measures the proportion of fully qualified trajectories, rather than the accuracy of individual frames or annotations.

\begin{table}[htbp]
    \centering
    \small
    \caption{Manual audit of automatically generated subtask annotations. A trajectory qualifies only if every subtask description and temporal boundary satisfies the audit criteria.}
    \label{tab:annotation_quality_audit}
    \begin{tabular}{lrrr}
        \toprule
        Dataset & Reviewed & Qualified & Pass rate \\
        \midrule
        DROID   & 55,864 & 18,682 & 33.44\% \\
        OXE     & 31,560 & 24,172 & 76.59\% \\
        Table30 & 2,962  & 1,286  & 43.42\% \\
        Others  & 2,079  & 1,576  & 75.81\% \\
        \midrule
        \textbf{Total}
        & \textbf{92,465}
        & \textbf{45,716}
        & \textbf{49.44\%} \\
        \bottomrule
    \end{tabular}
\end{table}

As shown in Table~\ref{tab:annotation_quality_audit}, 45,716 trajectories pass the audit, yielding an overall pass rate of 49.44\%.
Manual inspection identifies missing or underspecified task instructions in DROID and accumulated segmentation errors in longer Table30 trajectories as common sources of failure.
Because a single annotation error can cause an entire trajectory to fail this stringent audit, rejected trajectories may still contain substantial useful supervision.
Their diverse visual scenes and task--subtask structures provide valuable pretraining signals, broadening the range of tasks and environments encountered by the model.
We therefore use the broader automatically annotated corpus for pretraining and select approximately 15K qualified trajectories for refinement, combining broad task coverage with precise subtask supervision.

\subsection{Ablation of the Value Head}
\label{app:value_head_ablation}

We evaluate the contribution of the continuous value head on the ID and OOD splits of ProcVQA.
Using the same jointly trained checkpoint, we compare progress predictions from the value head with progress values parsed from the generated text.
The \emph{w/o Value Head} condition bypasses the value head only at inference; both conditions retain the same training history.
This comparison therefore evaluates the choice of progress readout at inference.
We report VOC for temporal ordering and EPR@50 for progress-score resolution.

\begin{table}[htbp]
    \centering
    \small
    \caption{Ablation of the value head on ProcVQA. Both conditions use the same trained checkpoint; the condition without the value head obtains progress scores from generated text. Higher values are better for both metrics.}
    \label{tab:value_head_ablation}
    \begin{tabular}{lcccc}
        \toprule
        & \multicolumn{2}{c}{ID}
        & \multicolumn{2}{c}{OOD} \\
        \cmidrule(r){2-3}
        \cmidrule(l){4-5}
        Variant
        & VOC $\uparrow$
        & EPR@50 $\uparrow$
        & VOC $\uparrow$
        & EPR@50 $\uparrow$ \\
        \midrule
        w/ Value Head
        & \textbf{0.8058}
        & \textbf{5.7489}
        & \textbf{0.7282}
        & \textbf{6.1152} \\
        w/o Value Head
        & 0.8031
        & 5.4400
        & 0.6843
        & 5.5457 \\
        \bottomrule
    \end{tabular}
\end{table}

Table~\ref{tab:value_head_ablation} shows that the value head improves both temporal ordering and prediction resolution.
On ID data, bypassing the value head has little effect on VOC but reduces EPR@50 by 5.4\%.
The effect is larger on OOD data, where VOC and EPR@50 decrease by 6.0\% and 9.3\%, respectively, relative to predictions from the value head.
These results support using the continuous value head for progress estimation, particularly on unseen tasks.

\subsection{Reward Modeling Cases}
\label{app:reward_case_study}

\begin{figure}[t]
    \centering
    \includegraphics[width=1\linewidth]{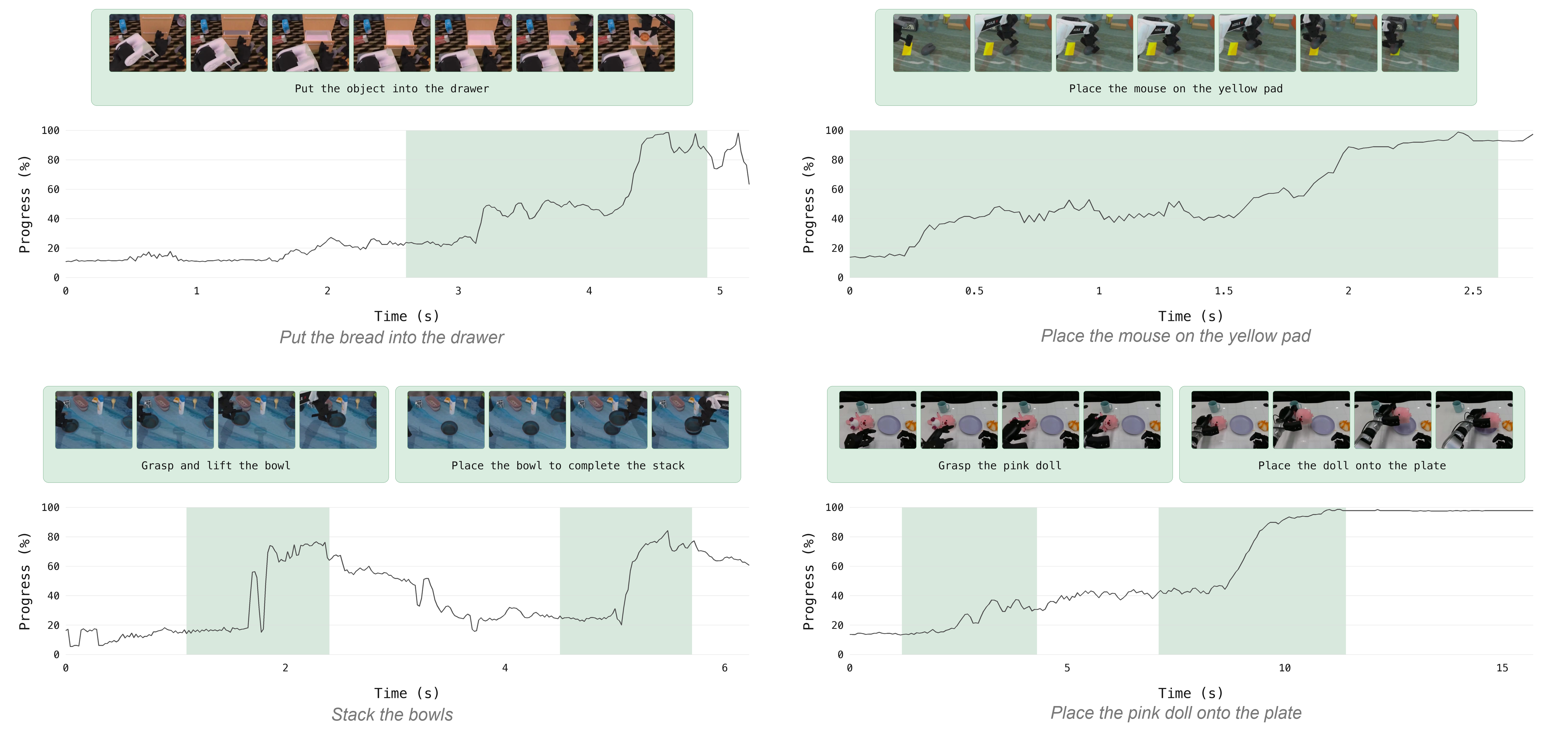}
    \caption{Zero-shot reward modeling cases on RoboMIND and RoboTwin. ProcVLM provides task-conditioned progress feedback across different manipulation tasks without task-specific adaptation.}
    \label{fig:zeroshot_cases}
\end{figure}

\textbf{Zero-shot cases.}
We first evaluate ProcVLM in a fully zero-shot setting on multiple tasks from RoboMIND \citep{robomind} and RoboTwin. 
As shown in Figure~\ref{fig:zeroshot_cases}, ProcVLM can identify the current action stage, reason about remaining steps, and produce task-conditioned progress feedback without task-specific demonstrations or reward-model adaptation. 
These cases illustrate the transferable progress perception learned from procedure-aware pretraining.

\begin{figure}[t]
    \centering
    \includegraphics[width=1\linewidth]{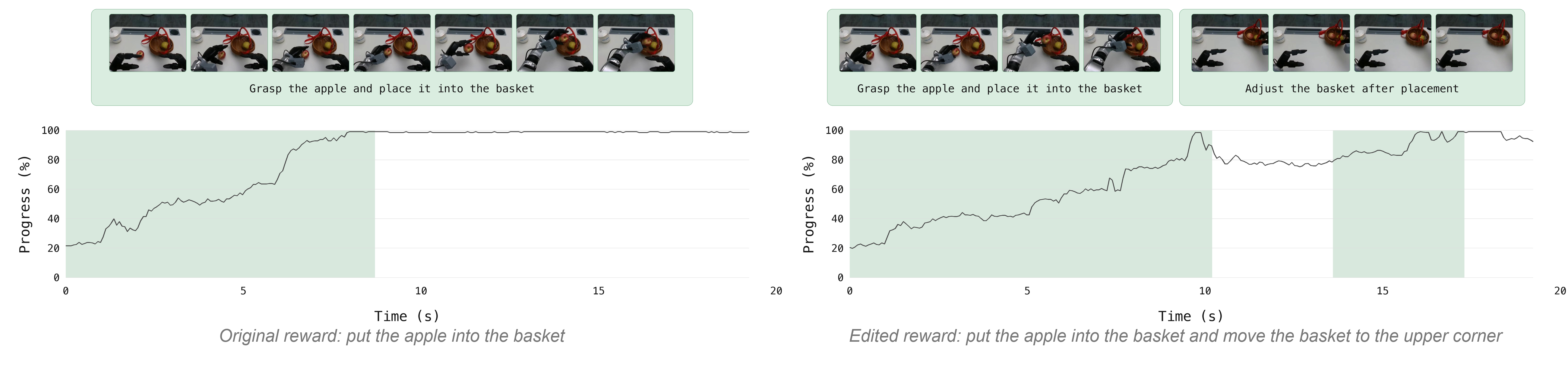}
    \caption{Zero-shot reward editing on the same video sequence. Left: reward for ``put the apple into the basket.'' Right: edited reward for ``put the apple into the basket and move the basket to the upper corner.''}
    \label{fig:reward_editing}
\end{figure}

\textbf{Zero-shot reward editing.}
We finally present a zero-shot reward editing case on a RoboMIND pick-and-place task. 
As shown in Figure~\ref{fig:reward_editing}, ProcVLM is applied to the same video under two task instructions: ``put the apple into the basket'' and ``put the apple into the basket and move the basket to the upper corner.'' 
The resulting reward curves change with the edited task description, showing that ProcVLM grounds reward prediction in task-conditioned subtask structure rather than only superficial visual progress.

\begin{figure}[t]
    \centering
    \includegraphics[width=1\linewidth]{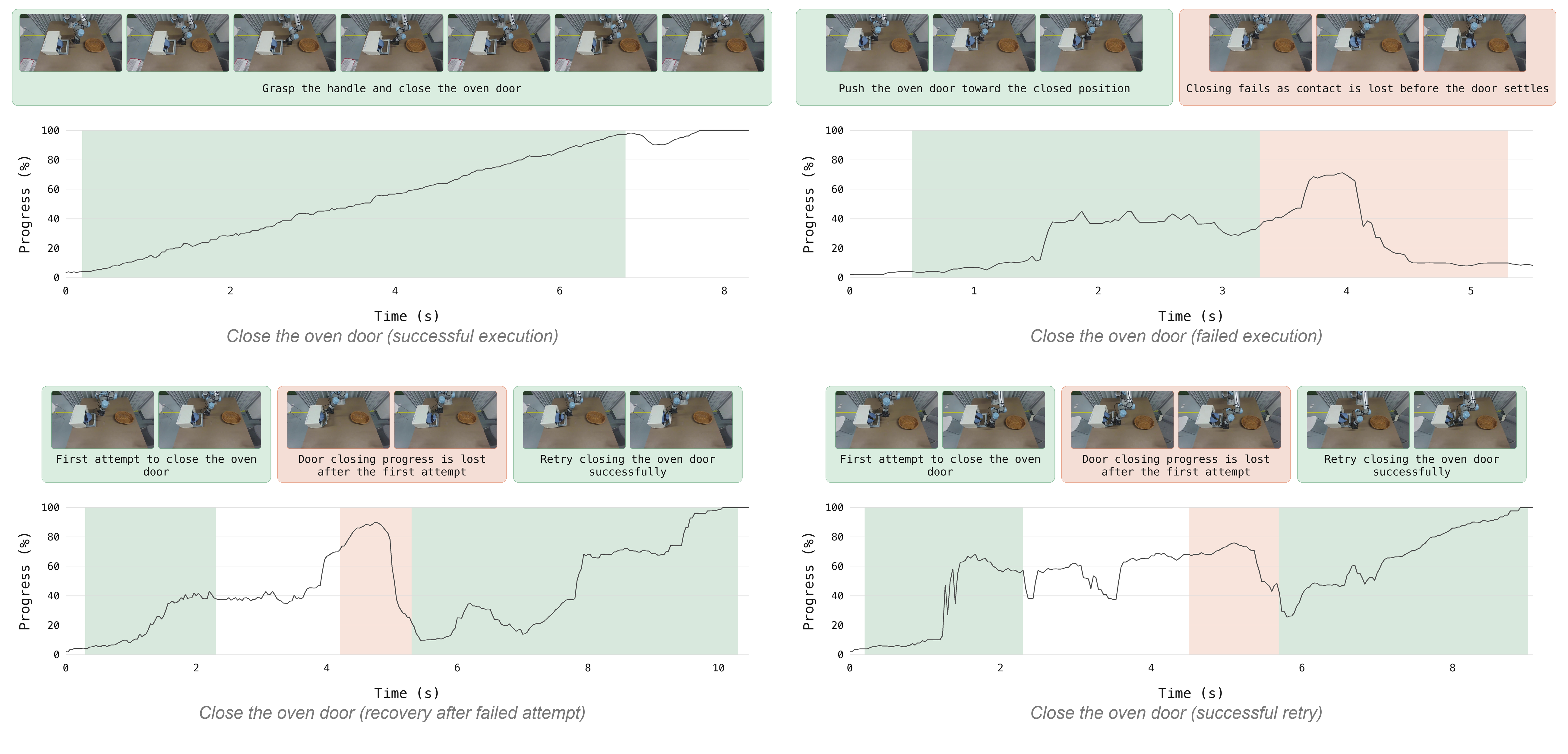}
    \caption{One-shot adaptation case for a close-oven task. ProcVLM adapts from one successful demonstration and evaluates different execution outcomes, including failures and retry-based recoveries.}
    \label{fig:oneshot_case_close_oven}
\end{figure}

\begin{figure}[t]
    \centering
    \includegraphics[width=1\linewidth]{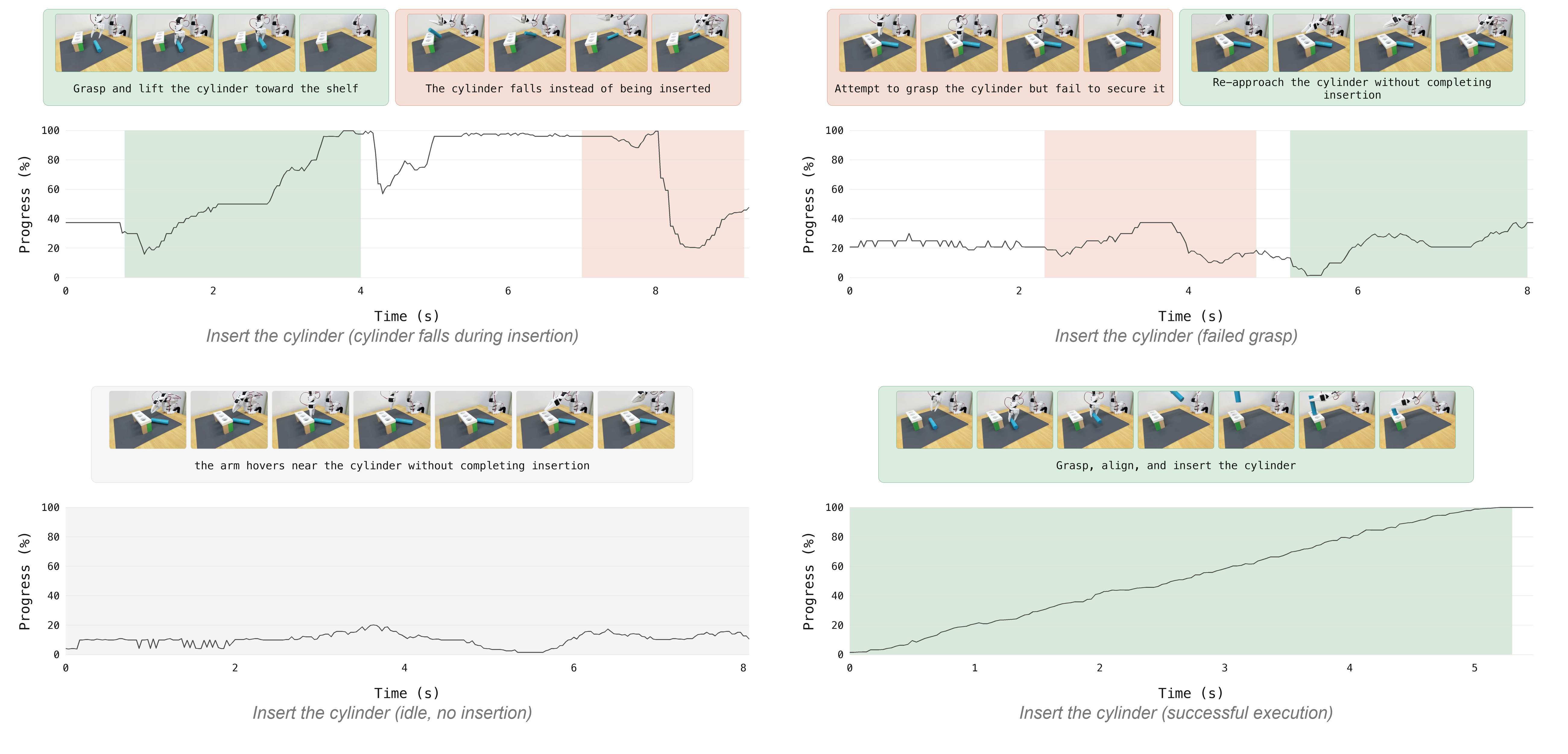}
    \caption{One-shot adaptation case for an insert-cylinder task. ProcVLM transfers the demonstrated task structure to executions with successful, failed, and idle behaviors.}
    \label{fig:oneshot_case_insert_cylinder}
\end{figure}

\begin{figure}[t]
    \centering
    \includegraphics[width=1\linewidth]{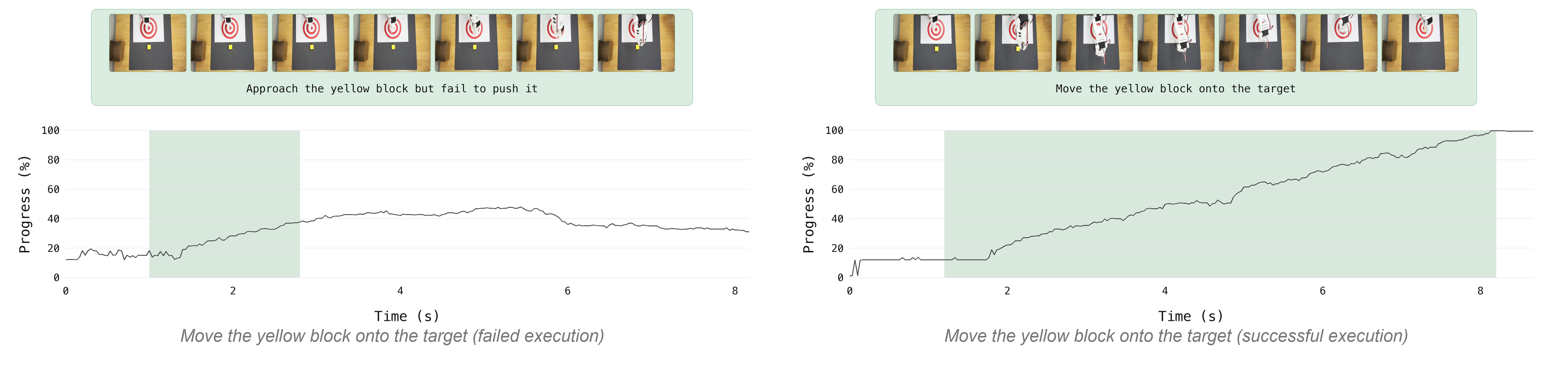}
    \caption{One-shot adaptation case for a move-target task. ProcVLM transfers the demonstrated task structure to successful and failed executions.}
    \label{fig:oneshot_case_move_target}
\end{figure}

\begin{figure}[t]
    \centering
    \includegraphics[width=1\linewidth]{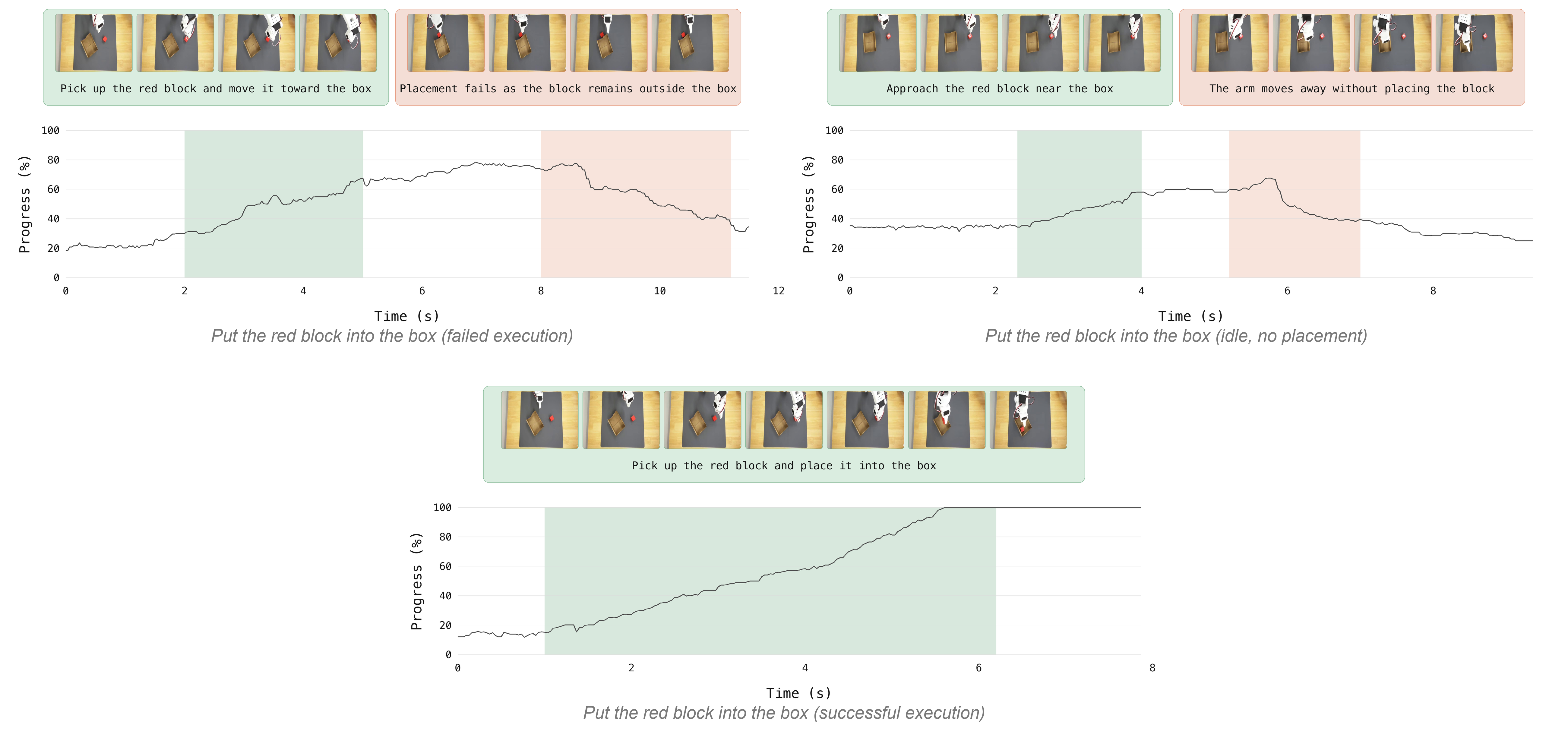}
    \caption{One-shot adaptation case for a put-in-box task. ProcVLM generalizes from one successful demonstration to evaluate successful and failed executions.}
    \label{fig:oneshot_put_in_box}
\end{figure}

\textbf{One-shot adaptation cases.}
We further evaluate ProcVLM under a one-shot adaptation setting, where only one successful demonstration is used for LoRA tuning in each scenario. 
As shown in Figures~\ref{fig:oneshot_case_close_oven}, \ref{fig:oneshot_case_insert_cylinder}, \ref{fig:oneshot_case_move_target}, and \ref{fig:oneshot_put_in_box}, ProcVLM quickly adapts to RoboMIND and RoboFAC scenarios containing successful, failed, and retry executions, providing meaningful progress feedback beyond the demonstrated success trajectory.

\newpage

\end{document}